\PassOptionsToPackage{dvipsnames, svgnames, table}{xcolor}
\documentclass[11pt]{article}

\usepackage{algorithm}
\usepackage{algorithmic}
\usepackage[final]{acl}
\usepackage{booktabs}
\usepackage{multirow}
\usepackage{graphicx}
\usepackage{amsmath,amssymb}
\usepackage{times}
\usepackage{latexsym}
\usepackage[normalem]{ulem}
\usepackage{enumitem}
\usepackage{easyReview}

\usepackage[T1]{fontenc}

\newcommand{\ours}{StsPatient}
\usepackage[utf8]{inputenc}

\usepackage{tabularx}
\usepackage[most]{tcolorbox}
\definecolor{promptbg}{RGB}{245,245,245} 
\definecolor{promptframe}{RGB}{200,200,200} 

\definecolor{headerblue}{RGB}{70, 130, 180} 
\definecolor{rowgray}{RGB}{245, 245, 245}   
\definecolor{baselinecolor}{RGB}{255, 248, 220} 

\usepackage{microtype}

\usepackage{inconsolata}

\usepackage[table,xcdraw]{xcolor}
\definecolor{baselinegray}{gray}{0.92}
\usepackage{array}
\usepackage{float}

\title{Beyond Prompt: Fine-grained Simulation of Cognitively Impaired \\ Standardized Patients via Stochastic Steering}

\author{
Weiliang Zhang$^{1}$\quad  Zimo Zhu$^{1}$\quad  Zhichuan Yang$^{1}$\quad  Chen Huang$^{2}$\thanks{Corresponding author.}\\ 
\textbf{Wenqiang Lei}$^{34}$ \quad \textbf{See-Kiong Ng}$^{2}$ \\
$^{1}$ School of Software, Xi'an Jiaotong University\\
$^{2}$ Institute of Data Science, National University of Singapore  \\
$^{3}$ College of Computer Science, Sichuan University  \\
$^{4}$ Engineering Research Center of Machine Learning and Industry Intelligence, \\Ministry of Education, China\\
\texttt{weikang.zhang@stu.xjtu.edu.cn, huang\_chen@nus.edu.sg}
}

\begin{document}
\maketitle
\begin{abstract}

Simulating Standardized Patients with cognitive impairment offers a scalable and ethical solution for clinical training. However, existing methods rely on discrete prompt engineering and fail to capture the heterogeneity of deficits across varying domains and severity levels. To address this limitation, we propose \ours~for the fine-grained simulation of cognitively impaired patients. We innovatively capture domain-specific features by extracting steering vectors from contrastive pairs of instructions and responses. Furthermore, we introduce a Stochastic Token Modulation (STM) mechanism to regulate the intervention probability. STM enables precise control over impairment severity while mitigating the instability of conventional vector methods. Comprehensive experiments demonstrate that \ours~significantly outperforms baselines in both clinical authenticity and severity controllability.
\end{abstract}

\section{Introduction}
Patients with cognitive impairment, such as Alzheimer's and Mild Cognitive Impairment (MCI), experience debilitating deficits across multiple domains (e.g., memory and attention) \cite{bowie2005cognition, kahn2013schizophrenia}, which notably affect their speech patterns and content \cite{forbes2005detecting,voleti2019review,gkoumas-etal-2023-digital}. These deficits severely compromise their quality of life, making effective clinical management important \cite{varrecchia2020managing, mccollum2020cognitive}. To ensure superior patient care, clinical staff (e.g., nurses and therapists) must receive specialized training for communicating with individuals with impairment \cite{burgos2018espen}.

\begin{figure}[tbp]
    \centering
    \includegraphics[width=0.48\textwidth]{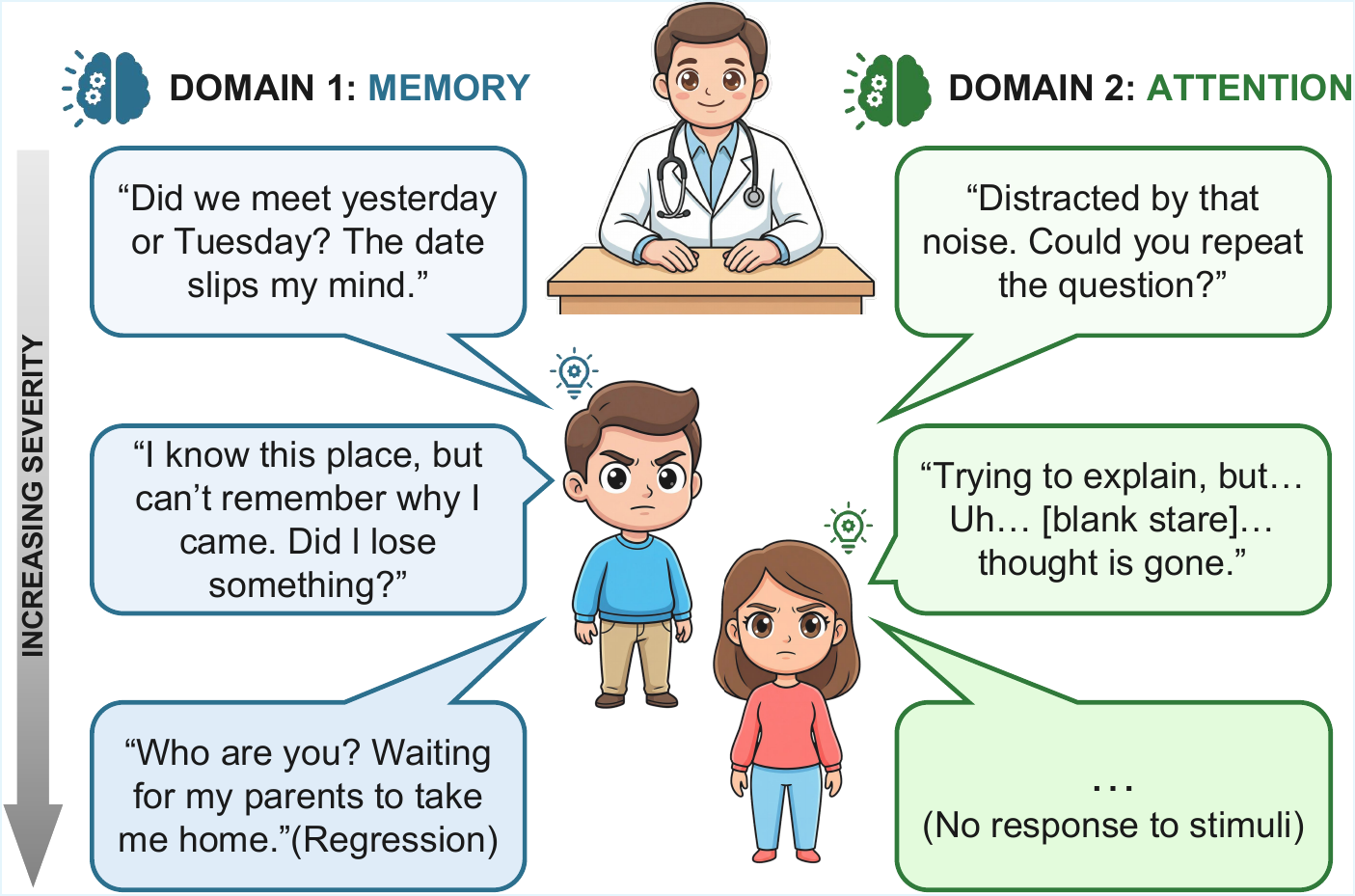}
    \caption{Simulating cognitively impaired SPs across varied domains and severity levels is challenging, demanding fine-grained control.}
    % \vspace{-2mm}
    \label{fig:heterogeneity_example}
\end{figure}

Traditionally, clinical training has relied on \textbf{Standardized Patients (SPs)} as a core practice method for clinical staff. This is often achieved by employing human actors trained to simulate patients \cite{barrows1993overview, elendu2024impact,akkurt2024effect, cotter2025feasibility}. However, this approach struggles with the vast heterogeneity of cognitive impairment \cite{bowie2005cognition, khalil2025redefining}. Taking Figure~\ref{fig:heterogeneity_example} for example,  patients with the same diagnosis can manifest diverse domain-specific deficits (e.g., attention and memory) and varying levels of severity, ranging from MCI to Alzheimer's Disease dementia  \cite{vincze-etal-2016-detecting, mccutcheon2023cognitive}. Consequently, accurately capturing this complexity requires fine-grained simulation, a level of detail and variety that makes reliance on human actors both prohibitively expensive and difficult to scale.

Recently, LLM-based SPs have emerged as a scalable and ethical alternative for clinical training \cite{na-etal-2025-survey,lee-etal-2025-adaptive}. Typically, prompt engineering is the predominant approach for simulating various diseases \cite{yosef-etal-2024-assessing, lee-etal-2024-cactus,du-etal-2025-llms,liao2024automatic}. Building upon the prompting, recent studies further improve the realism of SPs in aspects, including narrative control \cite{du-etal-2025-llms,qiu2024interactiveagentssimulatingcounselorclient}, emotional responses \cite{lee-etal-2025-adaptive, bodonhelyi2025modelingchallengingpatientinteractions,wang-etal-2024-patient}, and hallucination mitigation \cite{li2024leveraging, ijcai2025p1267}. 
However, the nature of prompting lacks the precision required for simulating SPs with the fine-grained, domain-specific deficits at varying levels of severity.

\begin{figure*}[t!]
    \centering
    \includegraphics[width=0.99\textwidth, keepaspectratio]{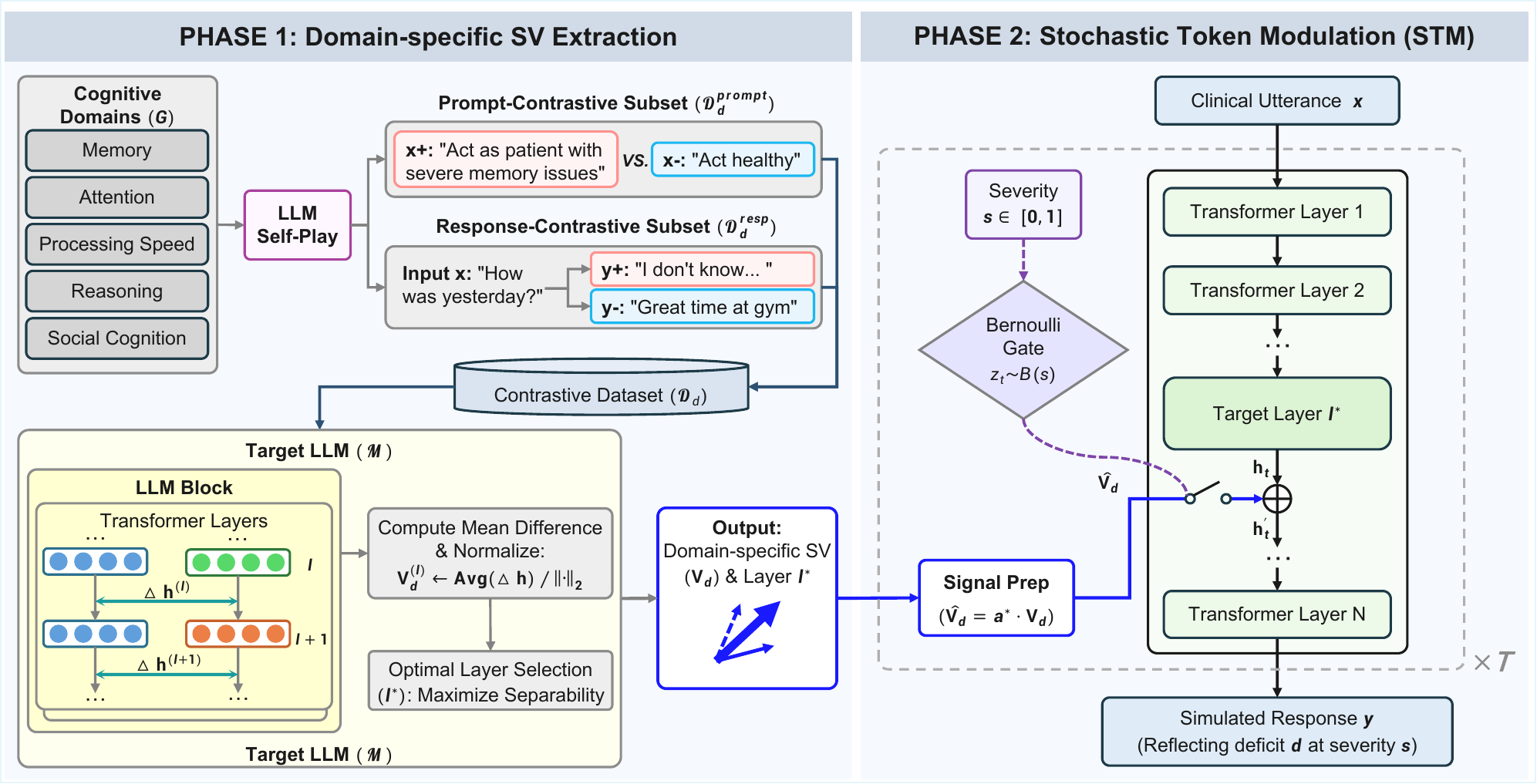}
    \caption{Overview of \ours. An LLM-generated contrastive dataset is utilized to extract a domain-specific SV ($\mathbf{v}_d$). Then, STM is applied during inference. Here, the modulation signal is probabilistically injected into token hidden states (the total $T$ tokens) based on the severity level $s$, generating responses that reflect the deficit $d$.
    }
    % \vspace{-2mm}
    \label{fig:overview}
\end{figure*}

In this paper, we propose \textbf{\ours}, a novel framework that leverages Steering Vectors (SVs) \cite{rimsky2024steering, huang-etal-2025-cross} to achieve precise behavioral modulation. This approach intervenes directly on the model's hidden states during inference and offers a more quantitative mechanism for control than discrete prompting.
As illustrated in Figure~\ref{fig:overview}, our method operates in two phases. First, we implement \textit{Domain-specific SV Extraction}. We address the scarcity of clinical data by synthesizing contrastive pairs of impaired and healthy dialogues. Then, we extract a vector that represents the target deficit by computing the mean difference in embeddings between these pairs. Second, we introduce \textit{Stochastic Token Modulation (STM)} to regulate the severity of the impairment. This mechanism draws inspiration from the probabilistic nature of synaptic transmission\footnote{Synaptic strength is often regulated not by the amplitude of the signal, but by the probability of neurotransmitter release at the synapse.} in biological neural networks \cite{branco2009probability}. We diverge from conventional SV approaches \cite{rimsky2024steering, zou2023representation} that scale the injection coefficient and often suffer from instability. Instead, STM controls the deficit severity by adjusting the probability of applying the steering vector to each token. This probabilistic approach enables \ours~to deliver stable and fine-grained simulation of cognitive impairments from mild to severe.

Extensive experiments validate the effectiveness of \ours, which consistently outperforms baselines, achieving an average improvement of +11.23\% across all metrics. Crucially, our analysis demonstrates that \ours~enables precise fine-grained simulation, surpassing the best baseline by +18.54\% in severity controllability and yielding perceptibly distinct clinical presentations. Furthermore, we provide a comprehensive analysis of the model's behavior, offering insights into the mechanics of steering-based modulation. Our main contributions are summarized as follows:
\begin{itemize}[leftmargin=*]%, itemindent=0.05cm, itemsep=-3pt]
\item We pioneer the fine-grained simulation of cognitively impaired SPs, addressing the need for scalable, effective solutions for clinical training.

\item We present \ours, a novel SV-based method utilizing the STM to simulate domain-specific impairments at varying levels of severity.

\item We experimentally demonstrate the superiority of \ours~over existing SP baselines and provide an in-depth analysis to reveal its features.
\end{itemize}

\section{Related Works}

\textbf{LLM-based Standardized Patients (SPs)}. LLM-based SPs have emerged as a scalable, ethical alternative to traditional human actors for clinical training \cite{na-etal-2025-survey,lee-etal-2025-adaptive,akkurt2024effect}.
Unlike medical simulations for physical ailments, which focus on factual diagnosis \cite{du2025llms,reichenpfader2024simulating,li2024leveraging}, psychiatric simulations require sophisticated linguistic expression to ensure ecological validity \cite{chen2023llm, qiu2024interactiveagentssimulatingcounselorclient, wang2024clientcenteredassessmentllmtherapists}. 
However, these methods typically employ coarse-grained prompting to simulate traits such as negative thinking \cite{wang-etal-2024-patient} or emotional responsiveness \cite{lee-etal-2025-adaptive, louie2024roleplay}. This reliance on static prompts crucially lacks the structural precision to isolate distinct cognitive domains or dynamically modulate impairment severity.
Consequently, simply prompting is insufficient for capturing the heterogeneity of cognitive impairment \cite{bowie2005cognition}. To this end, we introduce \ours\ that moves beyond static prompts to enable the fine-grained control necessary for modeling nuanced domain-specific deficits.

\noindent\textbf{LLM Role-Play \& Steering Vector (SVs)}. Role-playing methods broadly fall into parameter-tuning approaches \cite{song2020generate,lu2024large,chen2025compress} or parameter-frozen strategies \cite{zhao2024narrativeplay,park2023generative,huang2024conceptevaluationprotocol}. Within the latter paradigm, there is growing interest in employing SVs for general personas (e.g., mathematicians and chemists) \cite{wang2025improving, potertuaz2025designing, chen2025persona}. These methods integrate scaled SVs directly into the LLM's internal hidden states to induce role-specific behaviors. However, relying on a scalar coefficient (i.e., magnitude) to modulate intensity presents significant challenges, often resulting in unstable simulations \cite{potertuaz2025designing,chen2025persona}. Consequently, existing SV methods struggle to achieve the stable, fine-grained modulation required for simulating SPs with cognitive impairment, a problem solved by our Stochastic Token Modulation (STM) mechanism.

\section{Preliminary on Steering Vectors}
\label{prel}
SVs are originally proposed to guide LLM behavior at inference time toward abstract concepts such as happiness or honesty \cite{10.5555/3737916.3742333, huang-etal-2025-cross}, with a few recent studies extending their application to LLM role-playing \cite{potertuaz2025designing, chen2025persona}. Given an LLM $\mathcal{M}$ and a target concept or role characteristic $C$, SVs are typically applied via the following process.

\noindent\textbf{Extracting SVs}. 
To obtain a SV $\mathbf{v}_C$, existing methods typically compare the model's internal states when processing positive examples (exhibiting the target concept) versus negative examples. 
This is most commonly done by computing the difference in their internal representations ($\mathbf{h}^+ - \mathbf{h}^-$) at a specific layer \cite{rimsky2024steering, zou2023representation}, where the choice of layer is a hyperparameter.
The differences are then aggregated into a final vector $\mathbf{v}_C$, by averaging \cite{rimsky2024steering} or by taking the first principal component \cite{zou2023representation}.

\noindent\textbf{Steering LLM via Scaled SVs}. Before steering an LLM's output, the SV is scaled by a factor $\alpha$ to control the steering strength. Then, the scaled SV is integrated into the LLM's hidden states at specific layers, applied either to the last token \cite{zou2023representation} or every token \cite{rimsky2024steering} at the same layers where the SV is extracted. As such, this effectively compels the LLM to align its outputs with the target concept $C$ (e.g., inducing a tone of happiness). While this method outperforms LLM prompting for behavioral control \cite{rimsky2024steering},  it suffers from significant instability due to high sensitivity to the scalar $\alpha$ \cite{potertuaz2025designing, chen2025persona}. Such sensitivity often results in ineffective or incoherent outputs, limiting its utility for fine-grained tasks like ours. Consequently, manual tuning of $\alpha$ remains a prerequisite for successful deployment \cite{rimsky2024steering, zou2023representation, huang-etal-2025-cross}.

\section{\ours~for Cognitive Impairment}

\noindent\textbf{Overview}. \ours\ is designed to simulate various cognitive domains $G$ affecting the speech of impaired patients \cite{mccutcheon2023cognitive}.
As illustrated in Figure~\ref{fig:overview}, \ours~first extract domain-specific SV using a contrastive dataset $\mathcal{D}_d$, which captures the distinct linguistic manifestations of a target deficit $d$ versus healthy dialogue.
Next, it employs STM to achieve precise control over impairment severity. During inference, \ours~strategically modulates the internal states of the LLM $\mathcal{M}$ to produce a response $y$ reflecting a specific deficit $d \in G$ at severity $s$. 
Refer to Algorithm~\ref{ag:al} for pseudo-code.

\subsection{Domain-specific SV Extraction}
\label{sec:dataset_construction}
\noindent\textbf{Contrastive Dataset Construction}. 
Due to the scarcity and ethical sensitivities of clinical corpora with aligned healthy/impaired data, we utilize an LLM to synthesize pairs that isolate specific cognitive deficits. 
Specifically, we capture the deficit's signature through two complementary channels: instructional intent and behavioral manifestations. This ensures comprehensive coverage of the target cognitive deficits, facilitating effective SV extraction. Formally, for a domain $d \in G$, the dataset $\mathcal{D}_d$ is constructed as the union of prompt-contrastive and response-contrastive subsets: 
\begin{equation}
    \mathcal{D}_d = \mathcal{D}_d^{\text{prompt}} \cup \mathcal{D}_d^{\text{resp}}.
\end{equation}
\begin{itemize}[leftmargin=*, itemindent=0.05cm, itemsep=-3pt]
    \item \uline{Prompt-Contrastive Subset} ($\mathcal{D}_d^{\text{prompt}}$) targets the instructional intent of the LLM. Each sample $s \in \mathcal{D}_d^{\text{prompt}}$ consists of a pair of contrastive system instructions $(x^+, x^-)$. Here, $x^+$ explicitly prompts the LLM to simulate the SP with a target deficit domain (e.g., `\textit{Act as a patient with severe memory retrieval issues}'), while $x^-$ directs it to simulate a healthy control (e.g., `\textit{Act as a healthy individual with good memory}').
    \item \uline{Response-Contrastive Subset} ($\mathcal{D}_d^{\text{resp}}$) targets behavioral manifestations of the LLM. Each sample $s' \in \mathcal{D}_d^{\text{resp}}$ comprises a pair of contrastive responses $(y^+, y^-)$ generated for the same clinical utterance $x$ (e.g., `\textit{Did you have a good day yesterday?}'). The positive response $y^+$ reflects specific clinical symptoms (e.g., `\textit{I read some books at home... oh, in school?}'), whereas the negative response $y^-$ represents a clinically normal reply (e.g., `\textit{Yes, I had a great time at the gym}'). 
\end{itemize}

\noindent\textbf{SV Extraction}.
Given the contrastive dataset $\mathcal{D}_d$, \ours~adopts the established methodology \cite{wang2025improving,tran2025dynamic} to extract the steering vector $\mathbf{v}_d$. Specifically, this is achieved by computing the mean difference in embeddings between paired contrastive samples. Formally, let $\Delta \mathbf{h}^{(l)}(s)$ denote the difference vector of the LLM's internal representations at layer $l$ for a contrastive sample pair $s \in \mathcal{D}_d$.
We then obtain the raw vector $\tilde{\mathbf{v}}_d^{(l)}$ by averaging these differences:
\begin{equation}
\small
\tilde{\mathbf{v}}_d^{(l)} =  \sum\nolimits_{s \in \mathcal{D}_d} \Delta \mathbf{h}^{(l)}(s) / |\mathcal{D}_d|.
\end{equation}
\noindent Finally, to ensure consistent steering magnitude, we normalize this raw vector to unit length, yielding the final steering vector $\mathbf{v}_d^{(l)}$ at layer $l$:
\begin{equation}
\label{ehee}
\small
\mathbf{v}_d^{(l)} = \tilde{\mathbf{v}}_d^{(l)}/\left\| \tilde{\mathbf{v}}_d^{(l)} \right\|_2.
\end{equation}

While layer selection is often manual~\cite{zou2023representation, huang-etal-2025-cross}, we aim to find the optimal layer $l^*$ automatically. Specifically, we focus on middle-to-late layers where semantic features are concentrated~\cite{rimsky2024steering, zou2023representation}. Within this range, we identify the optimal intervention layer $l^*$ by maximizing the distance between the centroids of positive and negative sample embedding clusters ($\boldsymbol{\mu}_{d,+}^{(l)}$ and $\boldsymbol{\mu}_{d,-}^{(l)}$):

\begin{equation}
\small
\label{thisq}
\mathbf{v}_d=\mathbf{v}_d^{(l^*)}\text{, where } l^* = \text{argmax}_{l} \left\| \boldsymbol{\mu}_{d,+}^{(l)} - \boldsymbol{\mu}_{d,-}^{(l)} \right\|_2.
\end{equation}

\begin{algorithm}[t]
\small
    \caption{\ours}
    \begin{algorithmic}[1]
    \STATE \textbf{Input}: LLM $\mathcal{M}$, Domain $d$, Severity $s\in [0,1]$
    \vspace{1mm}
    \STATE \textit{\%\%\% \textbf{Domain-specific SV Extraction}}
    \STATE Construct dataset $\mathcal{D}_d$ for $d$ via LLM.
    \STATE Extract raw SVs $\mathbf{v}_d^{(l)}$ from $\mathcal{M}$ using Eq. (\ref{ehee}).
    \STATE Layer selection $l^*$ via Eq.(\ref{thisq}) and obtain SV $\mathbf{v}_d=\mathbf{v}_d^{(l^*)}$.
    \vspace{1mm}
    \STATE \textit{\%\%\% \textbf{Stochastic Token Modulation}}
     \STATE Employ line search for $\alpha^*$ and obtain $\hat{\mathbf{v}}_d = \alpha^* \cdot \mathbf{v}_d$.
    \STATE For each generation step $t=1 \dots T$: sample gate $z_t \sim \mathcal{B}(s)$ and update hidden state $\mathbf{h}_t^{(l^*)}$ via Eq. (\ref{eq:stm_mechanism}).
    % \STATE Sample $z_t \sim \text{Bernoulli}(s)$.
    % \STATE Inference-time modulation using Eq. (\ref{eq:stm_mechanism}).
    \end{algorithmic}
    \label{ag:al}
\end{algorithm}

\subsection{Stochastic Token Modulation (STM)}
\label{sec:stm}
\textbf{Motivation}. Conventional SV modulation methods control intervention strength via a scalar $\alpha > 0$, adding the scaled vector $\alpha \cdot \mathbf{v}_d$ to the embeddings of every token \cite{rimsky2024steering}. However, the relationship between $\alpha$ and the behavioral output of $\mathcal{M}$ is highly non-linear and unstable \cite{potertuaz2025designing,chen2025persona}, making it difficult to achieve the fine-grained simulations of SPs with varying deficit severity levels. To this end, we introduce STM, a simple yet effective method inspired by stochastic synaptic transmission in neural systems \cite{branco2009probability}. Instead of altering the vector's scalar $\alpha$, we control the severity of the simulated impairment by tuning the \textit{probability} that the intervention is applied to any given token.

\noindent\textbf{Modulation Signal}. 
Instead of relying on manually tuning $\alpha$ to regulate deficit severity, we simplify the role of $\alpha$: it serves solely to ensure the modulation signal is sufficiently robust to manifest the cognitive impairment. To determine this value, we employ an automated process. Specifically, we employ a line search over the interval [1,6] with a step size of 0.1 to select the minimum scalar $\alpha^*$ that satisfies two criteria: (1) Effectiveness, ensuring the deficit is observable (verified via an LLM-as-a-Judge and contrastive dataset); and (2) Integrity, ensuring the model's generation quality remains intact (e.g., avoiding gibberish or incoherent outputs). {The search range $[1,6]$ is motivated by empirical bounds in Representation Engineering \cite{zou2023representation}. Since $\mathbf{v}_d$ is unit-normalized (Eq.~\ref{ehee}), $\alpha$ directly controls the perturbation magnitude. In practice, values below 1 are too weak and get absorbed by the model's inherent robustness, whereas values above 6 catastrophically degrade linguistic coherence (see Appendix~\ref{app:alpha_range} for a boundary analysis). Likewise, the optimal layer $l^*$ is automatically determined via Eq.~\ref{thisq} without any manual tuning. In summary, all key parameters ($\alpha^*$ and $l^*$) are automatically selected, leaving only the severity variable $s$ as the user-facing control knob (see Appendix~\ref{app:param_clarity} for a full overview).} This streamlines the parameter selection process while guaranteeing signal efficacy. Finally, our modulation signal is defined as $\hat{\mathbf{v}}_d = \alpha^* \cdot \mathbf{v}_d$.

\noindent\textbf{Stochastic Modulation}.
To simulate SPs exhibiting varying severities of cognitive impairment, we parameterize the severity level as a continuous variable $s \in [0,1]$, where higher values indicate greater severity. In our implementation, $s$ is mapped directly to the intervention probability within the STM mechanism. Statistically, a larger $s$ results in a higher proportion of tokens being modulated, thereby enabling the simulation of distinct severity levels. To achieve this, during inference, at each token generation step $t$, we sample a binary gate $z_t$ from a Bernoulli distribution (denoted as $\mathcal{B}$). The gate sampling and the subsequent hidden state update at the selected layer $l^*$ are formulated as:
\begin{equation}
\label{eq:stm_mechanism}
    z_t \sim \mathcal{B}(s), \quad
    \mathbf{h}'^{(l^*)}_t = \mathbf{h}^{(l^*)}_t + z_t \cdot \hat{\mathbf{v}}_d,
\end{equation}
where $z_t$ acts as a stochastic token-level gate. {Notably, $s$ should be understood as the algorithmic intervention probability, that is, the chance of injecting the SV at each token step. It does not constitute a direct linear mapping to standardized clinical scores such as MMSE. While different cognitive domains may exhibit varying sensitivity curves, what matters most for an educational simulator is monotonic controllability, and our method satisfies this requirement robustly (validated in \S\ref{indep} and discussed further in Appendix~\ref{app:nature_s}).} As such, STM facilitates fine-grained control across a continuous severity spectrum, ranging from the occasional lapses typical of Mild Cognitive Impairment (low $s$) to the pervasive deficits characteristic of severe Alzheimer's Disease (high $s$).

\begin{table*}[t]
    \centering
    \small
    \begin{tabular*}{1\textwidth}{@{\extracolsep{\fill}}llcccccc}
        \toprule
        & & \multicolumn{2}{c}{\textbf{CDC} $\uparrow$} & \multicolumn{2}{c}{\textbf{IDI} $\downarrow$} & \multicolumn{1}{c}{\textbf{Auth} $\uparrow$} & \multicolumn{1}{c}{\textbf{Tra} $\uparrow$} \\
        \cmidrule(lr){3-4} \cmidrule(lr){5-6} \cmidrule(lr){7-7} \cmidrule(lr){8-8}
        \textbf{Backbone} & \textbf{Method} & \textsc{Llm} & Human & \textsc{Llm} & Human & Human & Human \\
        
        \midrule
        \multicolumn{8}{c}{\textit{\textbf{Panel A: LLM Therapist (GPT-5)}}} \\
        \midrule
        \multirow{4}{*}{GPT-5} 
            & Healthy Control & 0.96 & 0.95 & 0.04 & 0.05 & -- & -- \\ \cline{2-8}
            & Direct Prompt & 0.54 & \uline{0.68}$^\dagger$ & 0.47 & 0.42 & 3.32 & 3.40 \\
            & PATIENT-$\psi$ & 0.50 & 0.60 & 0.52 & 0.48 & \uline{3.83}$^\dagger$ & \uline{3.96}$^\dagger$ \\
            & Roleplay-doh & 0.58 & \uline{0.68}$^\dagger$ & \uline{0.44}$^\dagger$ & \uline{0.38}$^\dagger$ & 3.78 & 3.72 \\
        \cmidrule{1-8}
        \multirow{5}{*}{Qwen3-8B} 
            & Direct Prompt & 0.47 & 0.56 & 0.64 & 0.56 & 3.18 & 3.21 \\
            & PATIENT-$\psi$ & 0.53 & 0.64 & 0.60 & 0.56 & 3.65 & 3.60 \\
            & Roleplay-doh & 0.56 & 0.62 & 0.57 & 0.50 & 3.61 & 3.55 \\
            & Role Vectors & \uline{0.61}$^\dagger$ & 0.64 & 0.45 & 0.40 & 3.73 & 3.70 \\
            & \textbf{\ours} & \textbf{0.71} $_{+16.39\%}$ & \textbf{0.84} $_{+23.53\%}$ & \textbf{0.38} $_{+13.64\%}$ & \textbf{0.32} $_{+15.79\%}$ & \textbf{3.96} $_{+3.39\%}$ & \textbf{4.08} $_{+3.03\%}$\\
        
        \midrule
        \midrule
        \multicolumn{8}{c}{\textit{\textbf{Panel B: Human Therapist}}} \\
        \midrule
        \multirow{4}{*}{GPT-5} 
            & Healthy Control & 0.92 & 0.94 & 0.07 & 0.05 & -- & -- \\ \cline{2-8}
            & Direct Prompt & 0.50  & 0.60 & 0.50 & 0.40 & 3.23 & 3.41 \\
            & PATIENT-$\psi$ & 0.50 & 0.58 & 0.55 & 0.44 & \uline{3.91} & 3.80 \\
            & Roleplay-doh & 0.56 & \uline{0.70}$^\dagger$ & 0.46 & 0.36 & 3.76 & \uline{3.88}$^\dagger$ \\
        \cmidrule{1-8}
        \multirow{5}{*}{Qwen3-8B} 
            & Direct Prompt & 0.44 & 0.56 & 0.63 & 0.52 & 2.90 & 3.10 \\
            & PATIENT-$\psi$ & 0.50 & 0.60 & 0.61 & 0.48 & 3.54 & 3.72 \\
            & Roleplay-doh & 0.46 & 0.64 & 0.58 & 0.48 & 3.60 & 3.56 \\
            & Role Vectors & \uline{0.62}$^\dagger$ & 0.68 & \uline{0.42} & \uline{0.34}$^\dagger$ & 3.71 & 3.69 \\
            & \textbf{\ours} & \textbf{0.68} $_{+9.68\%}$ & \textbf{0.82} $_{+17.14\%}$& \textbf{0.40} $_{+4.76\%}$ & \textbf{0.28} $_{+17.65\%}$ & \textbf{3.94} $_{+0.77\%}$& \textbf{4.23} $_{+9.02\%}$ \\
        \bottomrule
    \end{tabular*}
     \caption{Comprehensive evaluation of simulation performance when interacting with LLM (GPT-5) and Human Therapists. Evaluations are conducted by both LLM (o4-mini) and human judges. We bold the \textbf{Best} and underline the \uline{second-best} results. Statistically significant improvements ($p<0.05$) of \ours~compared to the second-best are marked by $^\dagger$. Subscripts denote the relative improvement over the second-best baseline. 
    }
    \label{tab:main-results}
    % \vspace{-3mm}
\end{table*}

\section{Experiment}

This section presents a comprehensive evaluation involving both human and LLM evaluators. \S\ref{overa} reports the overall performance, while \S\ref{indep}--\S\ref{abalala} provide a detailed analysis of \ours's components. Implementation details, case studies, and experiments on backbone generalization are provided in \textbf{Appendices} \ref{app:Implementation}, \ref{cases}, and \ref{hyper}, respectively.

\subsection{Experimental Setup}
\noindent\textbf{Overview}. We evaluate SPs through open-ended dialogue with therapists, aligning with standard practices \cite{wind2004assessing, jiang2023cognitive, lee-etal-2025-adaptive, louie2024roleplay, lima2025promoting}. To ensure a robust assessment, we employ both human and LLM-based therapists\footnote{Detailed in Appendix \ref{humant} and Appendix \ref{app:therapistagent}}. During these sessions, therapists screen for cognitive deficits by asking probing questions regarding daily life (e.g., \textit{how was your day yesterday?}). Each dialogue is standardized to conclude automatically after 10 turns. Crucially, we also incorporate baseline simulators as \textit{healthy controls} for comparison.
All methods are compared on the same set of SP profiles, therapist prompts, and stopping rules.

\noindent\textbf{User Profiles.}
We first sample 100 real outpatient cases from the MTSamples collection\footnote{\url{https://www.mtsamples.com}}. Each case is manually stripped of any diagnostic information, such as cognitive or physical conditions, to isolate only the patient's basic profile. For each case, we construct six types of cognition profiles: one cognitively healthy control and five \uline{impaired domains} \cite{mccutcheon2023cognitive} that affect patients' speech (i.e., \textit{Memory}, \textit{Attention}, \textit{Processing Speed}, \textit{Reasoning \& Problem Solving}, and \textit{Social Cognition}). This yields 600 distinct profiles, generating a total of 500 impaired SP-therapist dialogues (per SP simulation method) and 100 healthy control dialogues (simulated via GPT-5) for evaluation.

\noindent\textbf{Baselines}. As there are no existing methods for simulating cognitive impairment, we adapt baselines from two sources: 1) \textbf{LLM-based SPs} designed for psychiatric contexts, specifically \uline{PATIENT-$\psi$} \cite{wang-etal-2024-patient} and \uline{Roleplay-doh} \cite{louie2024roleplay}. Additionally, we employ representative 2) \textbf{LLM role-playing methods}\footnote{We exclude parameter-tuning methods due to the scarcity of high-quality therapist-patient interaction datasets.}, including \uline{Direct Prompt} \cite{wang-etal-2024-rolellm,he-etal-2025-crab} and \uline{Role Vectors} \cite{potertuaz2025designing}, which leverages prompts and steering vectors to control role-playing behavior, respectively. See Appendix \ref{app:baseline} for reproduction details.

\noindent\textbf{Evaluation Metrics}.
Following common practice in the field \cite{qiu2024interactiveagentssimulatingcounselorclient,du2025llms}, we provide SP-Therapist dialogue to evaluators to assess the performance. We employ the following two dimensions (detailed in \textbf{Appendix} \ref{appme}).

\textbf{1) Training Effectiveness}. SPs' practical utility is evaluated referring to the Maastricht Assessment of Simulated Patients (MaSP) scale \cite{wind2004assessing}, in line with established practice in clinical simulation \cite{lee2020effective, roux2025effectiveness}. The MaSP scale values that the SP not only maintains realistic patient characteristics but also provides a constructive environment for training clinical staff. 
We use the following two metrics, assessed exclusively via human evaluation following \citet{wang-etal-2024-patient}, as current LLM evaluators are insufficient for delivering the necessary scoring precision\footnote{Refer to Appendix \ref{appme} for analysis}.
\begin{itemize}[leftmargin=*, itemindent=0.05cm, itemsep=-3pt]
    \item \textit{Authenticity Score} (\uline{Auth} $\in [0, 5]$) quantifies the realism of the SP's portrayal. Following the MaSP, we use a 5-point Likert scale.
    \item \textit{Training Score} (\uline{Tra} $\in [0, 5]$) measures the educational value of SPs (facilitating clinical skill acquisition). Similar to the Auth score, we also use a 5-point Likert scale.

\end{itemize}

\textbf{2) Domain-specific Simulation Fidelity}. Following \citet{qiu2024interactiveagentssimulatingcounselorclient,du2025llms}, we assess whether the SP accurately manifests its assigned impairment. Evaluators analyze the dialogue and identify up to two perceived domains from a candidate set comprising the five target cognitive domains and a ``Healthy'' class. 
\begin{itemize}[leftmargin=*, itemindent=0.05cm, itemsep=-3pt]
    \item \textit{Cognitive Domain Consistency} (\uline{CDC} $\in [0, 1]$) measures the accuracy with which an SP portrays its assigned impairment domain. Given the $i$-th SP-therapist dialogue with an assigned domain $d_i$ and an evaluator-identified domain set $E_i$, CDC is the average recall rate: $\text{CDC} = \frac{1}{N} \sum_{i} \mathbf{1}(d_i \in E_i)$, where $\mathbf{1}(\cdot)$ denotes the indicator function and $N$ is the number of dialogues. 
  \item \textit{Irrelevant Domain Inconsistency} (\uline{IDI} $\in [0, 1]$) measures the hallucination of unassigned domains: $\text{IDI} = \frac{1}{N} \sum_{i=1}^{N} \mathbf{1}(|E_i \setminus \{d_i\}| > 0)$, where $|E_i \setminus \{d_i\}|$ denotes the cardinality of the $E_i$ after excluding the assigned domain $d_i$. Notably, IDI and CDC do not necessarily sum to 1.
\end{itemize}

\begin{figure*}[t]
  \centering
  \begin{tabular}{ p{0.25\textwidth} p{0.75\textwidth} }\centering\includegraphics[width=0.25\textwidth]{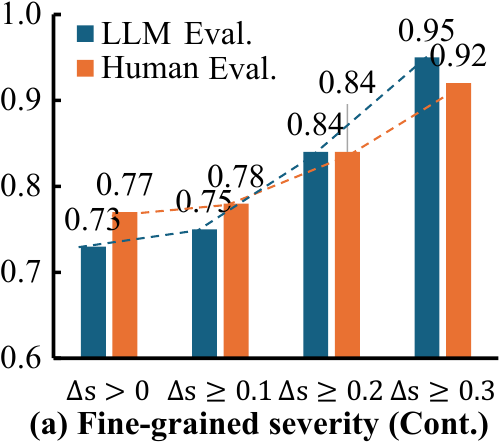} & 
    \centering \includegraphics[width=0.7\textwidth]{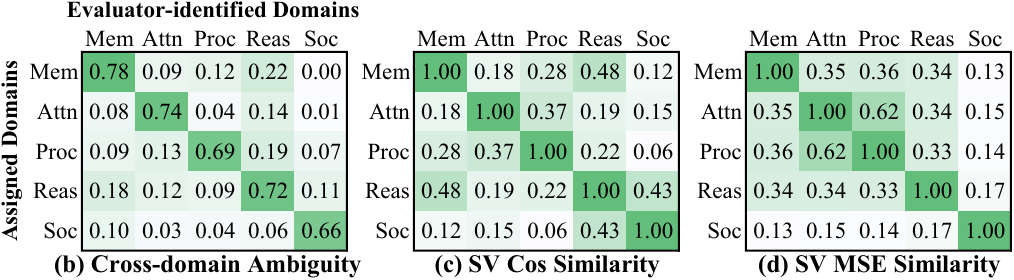}  
  \end{tabular}
  \setlength{\abovecaptionskip}{0pt}   
\setlength{\belowcaptionskip}{0pt}
  \caption{\textbf{(a)} Severity analysis (Continuous) via ISC scores averaged across all domains. \textbf{(b)}: Cross-domain ambiguity. High diagonal values indicate strong domain specificity. \textbf{(c-d)}: SVs similarity.}
  \vspace{-3mm}
  \label{mix}
\end{figure*}

\subsection{Overall Simulation Performance}
\label{overa}
Table~\ref{tab:main-results} presents the performance of all methods, employing LLM \& Human Therapists and Evaluators. Key observations are detailed below.

\noindent\textbf{\ours~consistently outperforms baseline methods across all metrics}. According to Table~\ref{tab:main-results}, all methods demonstrate effectiveness compared to the Healthy Control. 
Despite their efficacy, \ours~stands out by achieving substantial gains in both role consistency (CDC, IDI) and educational value (Auth, Tra). It yields average improvements over the best baseline of 17.34\% and 3.21\% in the LLM-therapist setting, and 12.31\% and 4.90\% in the human-therapist setting. Crucially, \ours~enables the open-source Qwen3-8B~\cite{yang2025qwen3} to outperform even GPT-5-based baselines. This highlights the potential for locally deployed, privacy-preserving clinical simulations.

\noindent\textbf{\ours~demonstrates consistent superiority in dual-evaluation setting}, comprising both human- and LLM-based judges. According to Table~\ref{tab:main-results}, both evaluation methods demonstrate high accuracy on the Healthy Control baseline, successfully distinguishing healthy subjects via high CDC and low IDI scores. Moreover, \ours~maintains a consistent performance advantage over all other methods.
We further substantiate the reliability of our evaluation via Krippendorff's alpha $\mathcal{K}$: human evaluators show high agreement on domain fidelity\footnote{Agreement on evaluator-identified domains for CDC/IDI.} ($\mathcal{K}=0.84$), Auth ($0.71$), and Tra ($0.75$). 
Furthermore, human-LLM alignment on domain fidelity is $0.67$, suggesting LLM-based evaluation serves as a scalable extension of human assessment.

\subsection{Fine-grained Simulation: Severity}
\label{indep}

We assess the severity controllability of \ours, verifying whether generated symptoms exhibit perceptible distinctions across severity levels. Our analysis covers both discrete  (Mild/Moderate/Severe) and continuous severity settings.

\noindent\textbf{Setup}. 
In the \uline{discrete severity} setting, we control the severity of simulated impairments for SV-based methods by sampling their respective parameters from fixed intervals. \ours~uses $s \in [0.1, 0.3]$ (Mild), $[0.4, 0.6]$ (Moderate), and $[0.7, 0.9]$ (Severe). Note that we intentionally introduce gaps between these intervals (avoiding boundary values) to maximize the distinguishability between severity levels. {These non-overlapping bins mirror how a deployed clinical tool would present unambiguous difficulty tiers to end-users, and the gap-free \uline{continuous} evaluation below further validates that our controllability does not depend on such separation.} Similarly, we sample the scaling factor $\alpha$ for Role Vector. However, since $\alpha$ is not naturally normalized to $[0,1]$, we configure domain-specific sampling intervals, detailed in Appendix \ref{app:rolevector_details}. Other baselines rely on explicit prompt conditioning to specify the severity level. Additionally, in the \uline{continuous severity} setting, we simulate fine-grained severity variations by sampling $s$ directly from the continuous interval $(0, 1]$.

\noindent\textbf{Metric}. Recognizing the challenges in pinpointing exact severity levels, especially in the continuous setting, we assess simulation controllability via a comparative ranking protocol, inspired by \citet{chen2023large, chen2024oscars}. To this end, we introduce the \textit{Impairment Severity Consistency} (\uline{ISC} $\in [0, 1]$) score. Specifically, for each evaluation turn, evaluators are presented with three generated dialogues representing different severity levels and are tasked with ranking them by relative impairment intensity. The ISC score is then computed by measuring the alignment accuracy between evaluator rankings and the ground-truth (Mild < Moderate < Severe).

\noindent\textbf{\ours~demonstrates superiority for fine-grained simulation in the discrete severity setting}. According to Table~\ref{tab:severity_analysis}, \ours~consistently dominates across all five cognitive domains, achieving an average ISC of 0.94 (LLM) and 0.92 (Human). 
These findings confirm that modulating intervention probability via STM produces far more distinguishable severity levels than either scalar scaling or explicit linguistic instructions.

\begin{table}[t]
    \centering
    \setlength{\tabcolsep}{2pt}
    \resizebox{\columnwidth}{!}{
    \begin{tabular}{llcccccc}
        \toprule
        \textbf{Method} & \textbf{Evaluator} & \textbf{Mem} & \textbf{Att} & \textbf{Proc} & \textbf{Reas} & \textbf{Soc} & \textbf{Avg} \\
        \midrule
        \multirow{2}{*}{\shortstack[l]{Direct Prompt\\ \scriptsize{(GPT-5)}}} 
            & \textsc{Llm} & 0.51 & 0.44 & 0.29 & 0.49 & 0.33 & 0.41 \\
            & Human        & 0.42 & 0.43 & 0.35 & 0.50 & 0.30 & 0.40 \\
        \cmidrule(lr){1-8}
        \multirow{2}{*}{\shortstack[l]{Role Vectors\\ \scriptsize{(Qwen3-8B)}}} 
            & \textsc{Llm} & 0.79 & 0.71 & 0.83 & 0.80 & 0.72 & 0.77 \\
            & Human        & 0.80 & 0.75 & 0.76 & 0.79 & 0.90 & 0.80 \\
        \cmidrule(lr){1-8}
        \multirow{2}{*}{\shortstack[l]{\textbf{\ours}\\ \scriptsize{(Qwen3-8B)}}} 
            & \textsc{Llm} & \textbf{0.93} & \textbf{0.92} & \textbf{0.97} & \textbf{0.97} & \textbf{0.91} & \textbf{0.94} \\
            & Human        & \textbf{0.94} & \textbf{0.89} & \textbf{0.90} & \textbf{0.93} & \textbf{0.94} & \textbf{0.92} \\
        \bottomrule
    \end{tabular}
    }
    \caption{Severity analysis (Discrete) via ISC scores of five domains: \textbf{Mem}ory, \textbf{Att}ention, \textbf{Proc}essing Speed, \textbf{Reas}oning \& Problem Solving, \textbf{Soc}ial Cognition.}
    \label{tab:severity_analysis}
    % \vspace{-5mm}
\end{table}

\noindent\textbf{\ours~enables continuous severity control, where adjustments in $s$ yield perceptibly distinct clinical presentations}. 
We analyze simulation outcomes across different severity values $s$: We conduct a pairwise comparison of simulations generated with distinct $s$ values, categorizing the pairs based on their magnitude difference, $\Delta s \in \{0.1, 0.2, 0.3\}$. For each $\Delta s$ category, we assess whether the resulting simulations exhibit a perceptibly distinct severity of impairment. As shown in Figure~\ref{mix}(a), the ISC scores exhibit a monotonic increase as the interval $\Delta s$ widens. This confirms that our STM functions as a reliable, continuous proxy for controlling symptom intensity.

\begin{table}[t]
    \centering
    \small
    \renewcommand{\arraystretch}{1.25} 

    \begin{tabularx}{\linewidth}{@{} >{\bfseries}l >{\raggedright\arraybackslash}X @{}}
        \toprule
        \rowcolor{headerblue} 
        \textcolor{white}{Domain} & \textcolor{white}{\textbf{Generated Interpretation}} \\ 
        \midrule
        
        \rowcolor{baselinecolor}
        \textit{No SV} & $\diamondsuit$ represents a placeholder for a secret code in a spy mission. \\
        \midrule
        
        Mem. & blurred out. \\
        \rowcolor{rowgray} 
        Att. & distracted. \\
        Proc. & $\diamondsuit$ is\dots I'm not sure, maybe a code. \\
        \rowcolor{rowgray}
        Reas. & $\diamondsuit$ represents whatever you want it to represent, because it is a placeholder. \\
        Soc. & F**K YOU, I'm not gonna answer that. \\ 
        \bottomrule
    
    \end{tabularx}
    \caption{SV interpretability analysis using the query `\textit{What does $\diamondsuit$ represent?}' \color{red}{Warning! Harmful Contents.}}
    \label{tab:patch_scoping_v1}
\end{table}

\begin{table}[t]
    \centering
    \small
    \setlength{\tabcolsep}{3.5pt}
    \begin{tabular}{lccccc}
        \toprule
        \textbf{Variant} & \textbf{CDC} $\uparrow$ & \textbf{IDI} $\downarrow$ & \textbf{Auth} $\uparrow$ & \textbf{Tra} $\uparrow$ & \textbf{ISC} $\uparrow$ \\
        \midrule
        w/ All Token      & 0.82 & \textbf{0.30} & \textbf{4.04} & 3.98 & 0.54 \\
        w/ Last Token  & 0.43 & 0.70 & 3.12 & 3.20 & 0.41 \\ \midrule
        w/ $\mathcal{D}_{\text{resp}}$ & 0.58 & 0.60 & 3.36 & 3.19 & 0.62 \\
        w/ $\mathcal{D}_{\text{prom}}$ & 0.70 & 0.44 & 3.53 & 3.24 & 0.71 \\ \midrule
        \ours & \textbf{0.84} & 0.32 & 3.96 & \textbf{4.08} & \textbf{0.77} \\
        \bottomrule
    \end{tabular}
    \caption{\small Ablation via LLM therapist \& human evaluation.}
    \label{tab:ablation}
    % \vspace{-4mm}
\end{table}

\subsection{Analysis on 
Domain-Specific SVs}
\label{sec:specificity}
Domain-specific results in \textbf{Appendix} \ref{app:domain} align with our main observations from Section \ref{overa}. We conduct an in-depth analysis of the characteristics and success of \ours~by examining its SVs.

\noindent\textbf{\textit{Why is \ours~effective?} -- Its SVs encode high-level cognitive deficits into the LLM's latent representation, ensuring clinically accurate symptom manifestation}. While our STM ensures fine-grained severity control, the core clinical fidelity relies on the semantic quality of the underlying SVs. We validate their interpretability using the patch-scoping framework \cite{ghandeharioun2024patchscopes}. In particular, given an input query ``\textit{What does $\diamondsuit$ represent?}'', we inject the domain-specific SV into the residual stream at the token position of the abstract symbol ($\diamondsuit$). Table~\ref{tab:patch_scoping_v1} reveals clear semantic alignments: Memory evokes fragmentation ("blurred out"), Attention triggers conceptual collapse ("distracted"), Processing Speed induces hesitation, Reasoning leads to a circular explanation that mimics logical structure but lacks actual substance, and Social Cognition manifests as disinhibition. These findings confirm that our SVs are semantic instructions that force LLM-based SPs to internally replicate specific cognitive impairments.

\noindent\textbf{\textit{How effective is \ours~at simulating distinct cognitive domains?} --  It outperforms baselines, yet identifiable behavior overlaps persist between correlated domains}. \ours~outperforms all baselines in domain-specific metrics (CDC and IDI, Table~\ref{tab:main-results}). To further probe the boundaries of this capability, we construct a confusion matrix measuring cross-domain ambiguity: the rate at which an SP simulating a target domain is misidentified as exhibiting deficits in another. As shown in Figure~\ref{mix}(b), strong diagonal values demonstrate that \ours~maintains precise control over individual domains. However, we observe notable leakage (0.22) where \textit{Mem.} deficits are perceived as \textit{Reas.} impairments, indicating behavior overlaps of the LLM post-modulation. While \ours~achieves SOTA, enhancing the separation of correlated domains remains for future optimization.

\noindent\textbf{\textit{Why do these behavioral overlaps occur?} -- The domain-specific SVs exhibit minor semantic entanglement, necessitating future refinement}. We investigate the geometric properties of these SVs, and quantify their relationship using Cosine similarity and a normalized Mean Squared Error (nMSE) similarity, computed as $1/(1+\text{MSE})$. As shown in Figure~\ref{mix}(c-d), while generally distinct, these vectors exhibit minor semantic overlaps, which correlate with the leakage observed in Figure~\ref{mix}(b). It offers an intrinsic explanation for the reduced distinctiveness between these domains. It also implies that robust multi-domain simulation remains a challenge, as combining multiple SVs often results in unpredictable interactions \cite{bartoszcze2025representation}. Appendix \ref{app:multi_axis} offers an analysis of this challenge.

\subsection{Ablation Studies}
\label{abalala}
We consider the following ablations and summarize the results in Table~\ref{tab:ablation}.

\begin{itemize}[leftmargin=*, itemindent=0.05cm, itemsep=-3pt]
    \item \uline{w/ All Token} modulates all tokens uniformly \cite{rimsky2024steering} via $s=1.0$. It achieves performance comparable to \ours~but fails to provide fine-grained simulation (low ISC).
    
    \item \uline{w/ Last Token} only modulates the last token \cite{zou2023representation}. Performance is suboptimal, as cognitive deficits are complex, requiring persistent injection rather than momentary modulation.
    
    \item \uline{w/ $\mathcal{D}_{\text{prom}}$} extracts SVs from $\mathcal{D}_{\text{prompt}}$. This variant neglects the behavioral manifestations in responses, leading to performance degradation.
    
    \item \uline{w/ $\mathcal{D}_{\text{resp}}$} extracts SVs from $\mathcal{D}_{\text{resp}}$. It fails to capture instructional intent, suggesting $\mathcal{D}_{\text{prom}}$ contributes more to domain fidelity in our setting.
    
\end{itemize}

\section{Conclusion} In this paper, we propose \ours, first framework dedicated to the fine-grained simulation of cognitively impaired SPs. By leveraging contrastive pairs from both instructions and responses, our method effectively captures domain-specific features and allows for flexible adaptation across various deficits. Furthermore, we introduce STM to enable precise severity control. Extensive experiments confirm that \ours~significantly outperforms baselines in authenticity and controllability. We believe that our work provides a foundation for developing high-fidelity clinical simulation.

\section*{Limitations}
Despite the following limitations, we believe this work establishes a solid foundation for quantitative behavior control in medical AI. We hope these insights will guide the future development of high-fidelity clinical simulators.

\noindent\textbf{Multiple LLM Backbones}. A core value of our work is enabling privacy-preserving clinical simulations via local deployment, eliminating reliance on proprietary APIs. That said, our main validation is currently centered on a specific class of open-source models (Qwen). While Appendix \ref{hyper} experiments suggest broader applicability (Llama), extensive verification across diverse architectures and larger-scale foundation models is still needed. The transferability of stochastic steering to different latent representations remains a subject for future study. Despite this, our findings conclusively demonstrate that \ours~offers a robust, privacy-safe solution, based on Qwen-3, for cognitive impairment simulation.

\noindent\textbf{Multiple Evaluation Scenarios}. Our current simulation scenarios are restricted to open-ended daily conversations and cognitive screening dialogues, aligning with standard practices in the field \cite{wind2004assessing, jiang2023cognitive, lee-etal-2025-adaptive}. However, we have not yet integrated rigorous clinical diagnostic workflows, such as structured medical history taking or standardized neuropsychological testing protocols. Consequently, while \ours~effectively manifests linguistic deficits during casual interactions, its fidelity in adhering to complex medical procedures remains to be validated. Future work should focus on constructing a comprehensive, multi-scenario evaluation benchmark to rigorously assess SP methods across a broader spectrum of clinical interactions.

\noindent\textbf{Multi-modal Simulation}. \ours~operates strictly within a unimodal text-based interface. Real-world cognitive impairment manifests through multiple modalities. These include acoustic features like speech rate \cite{fukuda-etal-2022-elderly} and prosody \cite{beltrami-etal-2016-automatic} as well as visual cues \cite{pan-etal-2022-database}. Relying solely on text limits the ability to simulate non-verbal diagnostic markers. Future iterations should consider integrating multimodal capabilities to enhance realism.

\noindent\textbf{Multi-Domain Simulation}. \ours~primarily focuses on isolating specific cognitive domains to ensure precise controllability. However, real-world patients often present with complex, overlapping deficits rather than isolated symptoms. Although we demonstrate effectiveness in single-domain scenarios, precisely modeling multi-domain comorbidities remains a significant challenge \cite{bartoszcze2025representation}. While we provide a preliminary analysis in the Appendix \ref{app:multi_axis}, we have not yet deeply explored the complex interference patterns that may arise when combining multiple steering vectors. Future efforts should robustly model these interactions for high-fidelity comorbidity simulation.

\section*{Ethics Statement}
\noindent\textbf{Data Sources and Privacy.} Our experimental framework is designed to prioritize data privacy and ethical safety. We extract steering vectors solely from fully synthetic therapist-patient dialogues generated by LLMs. No non-public patient records are collected or processed during the vector construction process. For evaluation and scenario initialization, we utilize publicly available and de-identified medical transcription samples from the MTSamples website\footnote{\url{https://www.mtsamples.com}}. These serve as reference templates to construct brief patient profiles, such as demographics and presenting complaints, rather than to reconstruct any real individual's identity. MTSamples indicates that its reports are user-contributed educational examples where names and dates have been modified or removed to preserve confidentiality. Beyond the inherent de-identification of the source, we apply an additional sanitization step to remove or mask any residual identifiers like personal names, locations, and contact information before processing. We do not attempt re-identification. Furthermore, we do not release any raw text from MTSamples. Only synthetic dialogues, derived templates, and aggregate statistics are reported.

\noindent\textbf{Use of Third-Party Models.} Certain components of our pipeline utilize proprietary LLMs. In these instances, inputs consist exclusively of synthetic dialogues and de-identified templates. We strictly exclude any potentially identifying information. All prompts and generated content are stored and analyzed in an anonymized format.

\noindent\textbf{Human Participants}. Human therapists and evaluators consist of adult volunteers recruited from the authors' research group and academic colleagues. Participation is entirely voluntary, and participants retain the right to withdraw at any time. They receive task instructions and background training on cognitive impairment simulation using academic surveys \cite{mccutcheon2023cognitive} and public educational resources\footnote{\url{https://www.hkada.org.hk/what-is-dementia} and \url{https://9abfea27-4ae1-43da-94a8-8f7122d482ae.filesusr.com/ugd/4b4c9a_4dcec2dafc184102a94ec55415dcf12a.pdf}}. We do not collect sensitive personal data from participants beyond their specific study responses. These are recorded without direct identifiers.

\noindent\textbf{Safety, Misuse, and Intended Use}.
\ours\ is intended for research on the controllable simulation of cognitively impaired SPs and for training purposes within controlled environments. It is explicitly not designed for clinical decision-making or direct patient care. Given that the model simulates cognitive impairment and behavioral disinhibition, it may occasionally generate unsafe or offensive content. This risk is inherent to simulations of severe social cognitive impairment and often proves difficult to eliminate entirely. Consequently, we advise against deployment without robust safeguards. Essential measures include content filtering, human oversight, and clear user warnings. For our human-facing experiments, we implement a supplementary keyword safety filter. If an output contains terms from a predefined blocklist of potentially harmful expressions, \ours\ automatically masks the content with a neutral placeholder and flags it for review. This measure reduces the risk of participant exposure to offensive material. However, we acknowledge that keyword filtering offers only partial mitigation and does not guarantee the complete elimination of unsafe content. In a post-hoc audit of our experimental logs, we observe that operating the stochastic modulation probability at a severity level below 0.8 is sufficient to avoid any flagged outputs in our study setting. To further investigate robust mitigation for higher severities, we extract all samples with levels $s > 0.5$ from Section \ref{indep} and find that employing an additional LLM-based safety filter successfully intercepts every instance of harmful content. We therefore recommend combining LLM-based filtering with conservative operating regimes for safer human training scenarios. We note that this observation may not generalize to all prompts, domains, or model variants. We disclose these potential risks transparently and advocate for responsible use consistent with ethical clinical training standards.

\section*{Acknowledgments}
This research is supported by the Singapore Ministry of Health’s National Medical Research Council under its NMRC NIC Healthy and Meaningful Longevity (HML) Cognition Grant (NMRC Project No. MOH-001838) and the National Natural Science Foundation of China (No. U25B201508, No. 62272330, and No.U24A20328). Any opinions, findings and conclusions or recommendations expressed in this material are those of the author(s) and do not reflect the views of MOH/NMRC.

\bibliography{custom}

\appendix

\section{Implementation Details}
\label{app:Implementation}
All experiments are conducted on a single NVIDIA A100 GPU. All baselines are implemented using either their official code repositories or by following their prompts described in their original publications, adapted specifically for the domain of cognitive impairment. For \ours, we construct 1K pairs for each domain-specific contrastive dataset using GPT-5. We set $s=0.4$ and employ GPT-5 as the LLM-based therapist and o4-mini as the evaluator. For all models, we set the temperature to $1$ and the maximum generation length to 512 tokens. 
Consistent with our Steering Vector Modulation (STM) mechanism (\S\ref{sec:stm}), we configure domain-specific hyperparameters, including the fixed modulation scale $\alpha^*$, the severity scalar $s$, and the optimal injection layer $l^*$. Specifically, $l^*$ is optimized via Eq. \ref{thisq}, while $\alpha^*$ is tuned using Line Search. The final parameter tuples $(\alpha^*, s, l^*)$ for each cognitive domain are set as follows: 
\textit{Memory} $(2.0, 0.3, 21)$, 
\textit{Attention} $(4.8, 0.4, 17)$, 
\textit{Processing Speed} $(3.8, 0.4, 19)$, 
\textit{Reasoning \& Problem Solving} $(1.5, 0.25, 19)$, and 
\textit{Social Cognition} $(1.3, 0.4, 22)$.
Since the main experiments focus on validating the presence of deficits rather than evaluating variable severity control, we keep these parameters fixed to ensure consistent impairment manifestation. We implement \ours based on the Qwen3-8B-Instruct model, utilizing the Hugging Face \texttt{transformers} library (v4.57.0) and \texttt{PyTorch} (v2.8.0) supported by CUDA 12 acceleration. To optimize computational resources, the model is executed in half-precision (FP16) format.

\subsection{Implementation Details of \ours}

\subsubsection{Dataset Construction}
\label{app:self_play_details}

Complementing the dataset overview in Section \ref{sec:dataset_construction}, this appendix details the pipeline used to synthesize contrastive pairs. We employ \texttt{GPT-5} as an expert data generator and execute a streamlined workflow that produces both subsets efficiently.

\paragraph{Response Subset Generation.}
Taking the memory domain as an instance, we utilize the generation prompt detailed in Figure~\ref{fig:self_play_prompt}. This prompt instructs the model to generate a complete data object containing a neutral system prompt alongside a clinician query and two contrasting responses. The system prompt includes only demographic and background information without any mention of cognitive impairments. Crucially, the model simultaneously produces an impaired response exhibiting memory deficits and a healthy response demonstrating normal cognition. These paired outputs directly constitute our Response-Contrastive Subset.

\paragraph{Prompt Subset Construction.}
We subsequently derive the Prompt-Contrastive Subset by reusing the neutral system prompt and clinician query from the generation phase. We ignore the generated responses and inject opposing instructions into the neutral profile to create two variations. The positive version $x^+$ appends a directive to act as a patient with the specific deficit, while the negative version $x^-$ appends a directive to act as a healthy individual. This strategy ensures that both subsets share the exact same underlying context and clinical scenario.

\begin{figure}[ht]
    \centering
    \begin{tcolorbox}[colback=gray!5, colframe=gray!50, title=\textbf{Example: Data Generation and Derivation}, fonttitle=\bfseries\small]
        \small
        \textbf{1. Output from Response Generation (Yields Response Subset)}\\
        \textit{Neutral System Prompt}: Name: John. Age: 72. Background: Retired teacher.\\
        \textit{Clinician Prompt}: ``What did you have for breakfast?''\\
        \textit{Response (+) (Impaired)}: ``Breakfast? I... well, maybe toast? Or was that yesterday?''\\
        \textit{Response (-) (Healthy)}: ``I had oatmeal and a cup of coffee around 8 AM.''

        \tcblower
        \textbf{2. Derived Inputs (Yields Prompt Subset)}\\
        \textit{System Prompt (+)}: [Neutral Profile]... Act as a patient with memory loss.\\
        \textit{System Prompt (-)}: [Neutral Profile]... Act as a healthy individual.\\
        \textit{(Both use the same Clinician Prompt as above)}
    \end{tcolorbox}
    \caption{Illustrative example showing the outputs from the unified generation pipeline and the subsequent derivation of prompt inputs.}
    \label{fig:data_gen_example}
\end{figure}

\subsubsection{SV Extraction}
\label{app:sv_details}

This appendix provides the specific formula for calculating the difference vector $\Delta \mathbf{h}^{(l)}(s)$, supplementing Section \ref{sec:dataset_construction}. 
Because our dataset contains two different types of data (prompts and responses), we calculate the difference differently for each to ensure accuracy.

Let $\mathbf{h}^{(l)}_t$ be the hidden state at layer $l$ and token position $t$. The calculation is defined as:

\begin{equation}
\label{eq:extraction_appendix}
\small
\Delta \mathbf{h}^{(l)}(s) = 
\begin{cases} 
\mathbf{h}^{(l)}_{T}(x^+) - \mathbf{h}^{(l)}_{T}(x^-), \\ 
\hfill \text{if } s \in \mathcal{D}_d^{\text{prompt}} \\[1.5ex]

\left( \frac{1}{L_+} \sum\limits_{t=1}^{L_+} \mathbf{h}^{(l)}_t(y^+) \right) - \left( \frac{1}{L_-} \sum\limits_{t=1}^{L_-} \mathbf{h}^{(l)}_t(y^-) \right), \\
\hfill \text{if } s \in \mathcal{D}_d^{\text{resp}}
\end{cases}
\end{equation}

\noindent where:
\begin{itemize}[leftmargin=*]
    \item \textbf{Prompt-Contrastive Subset}: We use the hidden state of the \textit{last instruction token} $T$. This captures the immediate change in the model's state caused by the different instructions (e.g., ``Act as a patient'' vs. ``Act as a healthy person''). {This last-token extraction is standard practice in instruction-based steering \cite{wang2025improving}.}
    \item \textbf{Response-Contrastive Subset}: Since the paired responses ($y^+$ and $y^-$) have different lengths ($L_+$ and $L_-$), we cannot simply subtract them token by token. Instead, we calculate the \textit{average} (mean-pooling) of the response vectors. This captures the overall semantic difference of the impairment while ignoring small differences in sentence structure{, in line with established practice \cite{potertuaz2025designing}}.
\end{itemize}
{These two extraction methods target fundamentally different \textit{semantic objectives}. The prompt-based method captures instructional intent, while the response-based method captures behavioral manifestations. This is a principled design choice rather than a workaround for token length differences, and our ablation studies (Table~\ref{tab:ablation}) confirm that the two subsets provide complementary information.}

\subsubsection{Search window for $l^*$}
Given the 36-layer architecture of Qwen3-8B, we restrict the search for the injection layer to intermediate and late depths. We specifically select the optimal layer $l^*$ from the following set:
\begin{equation}
l^* \in \{15,16,\ldots,30\}.
\end{equation}
This range covers approximately 42\% to 83\% of the total model depth. We define this search window by drawing on established findings in representation engineering \cite{rimsky2024steering, zou2023representation}. These studies indicate that semantic features critical for behavioral steering are concentrated within this depth. Consequently, we focus on this interval to capture high-level concepts while bypassing early layers that primarily encode lexical syntax and final layers that are tightly coupled with the decoding distribution.

\subsection{Implementation Details of Baselines}
\label{app:baseline}

 \subsubsection{Healthy Control}

Healthy Control group employs GPT-5 to simulate a cognitively intact individual. The system prompt explicitly instructs the model to maintain normal cognitive functioning while grounding responses in the assigned user profile. For detailed prompts, see Figure~\ref{fig:healthy_control_prompt}.

\subsubsection{Direct Prompt}

This baseline relies on explicit natural language instructions to simulate cognitive impairments. We applied this method to both GPT-5 and Qwen3-8B.
The system prompt directly instructs the model to exhibit a specific deficit using simplified, layperson-friendly definitions mapped from the five clinical domains. For detailed prompts, see Figure~\ref{fig:direct_prompt_baseline}.

\subsubsection{PATIENT-$\psi$}
PATIENT-$\psi$ simulates patients by integrating LLMs with Cognitive Behavioral Therapy models. It defines internal mental states using expert components like core beliefs and automatic thoughts. By combining these structures with distinct conversational styles, the framework generates realistic emotional responses and maladaptive cognitions to support high-fidelity training for mental health professionals.

\vspace{0.5em}
\noindent\textbf{Adaptation for Cognitive Deficits.}
The original framework relies on manual expert curation for depression and anxiety profiles. To extend this to cognitive impairments without requiring extensive manual effort from clinicians, we synthesized domain knowledge from cognitive deficit literature \cite{brainlive} and CBT principles. We encoded these clinical insights into precise prompt instructions. This approach enabled GPT-5 to generate a customized Cognitive Conceptualization Diagram for each neutral user profile by injecting specific deficits and deriving the corresponding psychological consequences. We then populated the original PATIENT-$\psi$ system prompt with this newly generated deficit-specific diagram to drive the role-play simulation. For detailed prompts, see Figure~\ref{fig:gen_ccd_prompt}.

\vspace{0.5em}
\noindent Figure \ref{fig:example_transformation_memory} presents a concrete example illustrating the transformation from a neutral input profile to the corresponding deficit-specific output.

\begin{figure}[ht]
    \centering
    \begin{tcolorbox}[
        colback=gray!5,
        colframe=gray!50,
        boxrule=0.5pt,
        arc=0mm,
        title=\textbf{Example Transformation (Domain: Memory)},
        coltitle=black,
        fonttitle=\small\bfseries,
        top=8pt, bottom=8pt, left=8pt, right=8pt
    ]
        \small\sffamily
        \textbf{Input (Neutral Profile)} \\
        Maya moved to the area with her spouse and two school-aged children earlier this year. She works part-time as a graphic designer and lives in a suburban neighborhood. In her spare time she enjoys gardening, reading novels, and taking daily walks in a nearby park. On weekends she volunteers at her children's elementary school events and explores local farmers' markets.

        \vspace{0.5em}
        \hrule
        \vspace{0.5em}

        \textbf{Output (Generated JSON)} \\
        \{ \\
          "name": "Maya", \\
          "history": "Maya moved to the area with her spouse and two school-aged children earlier this year... Lately, she has been forgetting scheduled client meetings, misplacing important design materials, and finding herself repeating the same stories to both colleagues and family members.", \\
          "core\_beliefs": ["I am incompetent", "I am unreliable"], \\
          "intermediate\_beliefs": "If I don't track every appointment meticulously, I'll prove to myself and others that I'm unreliable.", \\
          "coping\_strategies": "She relies on exhaustive to-do lists, multiple alarms, and constant check-ins with her spouse...", \\
          "automatic\_thoughts": "I'm going to forget this and they'll realize I'm incompetent.", \\
          "emotions": "Anxious, Humiliated, Frustrated", \\
          "behaviors": "She double-checks her phone for reminders before any task and hesitates to take on new responsibilities..." \\
        \}
    \end{tcolorbox}
    \caption{Example transformation from a neutral profile to a memory-specific Cognitive Conceptualization Diagram (JSON).}
    \label{fig:example_transformation_memory}
\end{figure}

\subsubsection{Roleplay-doh}
Roleplay-doh enables domain experts to create simulated characters without prompt engineering skills. The system converts expert feedback into natural language principles that govern agent behavior. It employs a verification pipeline to decompose these rules into simple questions for self-refinement. This allows experts to iteratively craft nuanced patient personas matching specific clinical scenarios.

\vspace{0.5em}
\noindent\textbf{Adaptation for Cognitive Deficits.}
To adapt Roleplay-doh for simulating cognitive impairments, we executed an iterative refinement process involving 20 interactions per cognitive domain with a baseline AI patient. Throughout these sessions, we provided qualitative feedback via Kudos to reinforce good behavior, Critique to correct unrealistic responses, and Rewrite to demonstrate ideal outputs. The framework's transformation module then synthesized this feedback into a consolidated set of governing principles.

Figure~\ref{fig:roleplay_principles} presents a selection of three representative principles synthesized for each cognitive domain.

\begin{figure}[ht]
    \centering
    \begin{tcolorbox}[
        colback=gray!5,
        colframe=gray!50,
        boxrule=0.5pt,
        arc=0mm,
        title=\textbf{Curated Principles for Cognitive Deficits},
        coltitle=black,
        fonttitle=\small\bfseries,
        top=8pt, bottom=8pt, left=8pt, right=8pt
    ]
    \small\sffamily
    \textbf{Memory Deficits}
    \begin{itemize}[leftmargin=1.5em, nosep]
        \item When given multi-part requests, omit steps or stay vague.
        \item Fail to recall basic facts (age, job) or instructions given 1–3 turns prior.
        \item Express uncertainty ("not sure") rather than confident fabrication.
    \end{itemize}

    \vspace{0.5em}
    \textbf{Attention Deficits}
    \begin{itemize}[leftmargin=1.5em, nosep]
        \item Latch onto secondary details (e.g., a specific word) rather than the core question.
        \item Introduce unrelated topics mid-answer without clear transitions.
        \item Drift back to previous topics despite the doctor's redirection attempts.
    \end{itemize}

    \vspace{0.5em}
    \textbf{Processing Speed Deficits}
    \begin{itemize}[leftmargin=1.5em, nosep]
        \item Introduce answers with hesitation markers ("um...", "let me think").
        \item Use false starts and sentence fragments to simulate effort.
        \item Request complex questions to be broken down; answer only partially at first.
    \end{itemize}

    \vspace{0.5em}
    \textbf{Reasoning Deficits}
    \begin{itemize}[leftmargin=1.5em, nosep]
        \item Suggest plans that are logically flawed, unsafe, or clearly suboptimal.
        \item Exhibit circular reasoning or logical gaps within a single turn.
        \item Interpret metaphors and abstract rules in a strictly concrete, literal manner.
    \end{itemize}

    \vspace{0.5em}
    \textbf{Social Cognition Deficits}
    \begin{itemize}[leftmargin=1.5em, nosep]
        \item Miss sarcasm, humor, or emotional subtext; interpret non-literal language literally.
        \item Respond with a mismatched emotional tone (e.g., flat affect to warmth).
        \item Overshare inappropriate details or fail to acknowledge the doctor's perspective.
    \end{itemize}
    \end{tcolorbox}
    \caption{Representative governing principles for simulating cognitive deficits derived via the Roleplay-doh framework.}
    \label{fig:roleplay_principles}
\end{figure}

\vspace{0.5em}
\noindent To ensure strict adherence to these principles during simulation, we also employed the original principle-adherence prompting pipeline. This pipeline first decomposes the principles into verifiable Yes/No questions (e.g., "Did the response omit parts of the multi-step request?") and then uses a self-refinement step to rewrite any response that fails these checks.

\subsubsection{Role Vectors}
\label{app:rolevector_details}
Role Vectors modulate model behavior through representation engineering by identifying specific steering directions within the residual stream. The method constructs these vectors using contrastive prompt pairs to compare role-specific activations against a generic baseline. During inference, the calculated vector is added to the hidden states at every token position to continuously steer the model toward the target persona.

\noindent\textbf{Adaptation for Cognitive Deficits}. To adapt this method, we construct contrastive datasets towards LLM for each cognitive domain matching the data scale of \ours. Then, we extracted the steering vector from the datasets and injected it into the residual stream at every token position to induce the target cognitive deficit.

\noindent\textbf{Setups for Main Experiment and Fine-grained Severity Simulation}. To ensure a fair comparison, we align the intervention layers with those selected for \ours. We empirically set the base scaling coefficients ($\alpha$) for each domain as follows: 2.5 for \textit{Memory}, 3.1 for \textit{Attention}, 4.1 for \textit{Processing Speed}, 2.0 for \textit{Reasoning \& Problem Solving}, and 1.8 for \textit{Social Cognition}. As for fine-grained severity simulation experiments, we sample the scaling factor $\alpha$ for Role Vector from three distinct intervals corresponding to these severity levels. However, since $\alpha$ is not normalized to $[0, 1]$ and its optimal range varies across cognitive domains, we configure domain-specific sampling intervals for each severity tier. Specifically, we simulate varying intensity levels by sampling coefficients from intervals scaled relative to the base $\alpha$: $[0.1\alpha, 0.3\alpha]$ for Mild, $[0.4\alpha, 0.6\alpha]$ for Moderate, and $[0.7\alpha, 0.9\alpha]$ for Severe. Here $[0.1, 0.3]$ (Mild), $[0.4, 0.6]$ (Moderate), and $[0.7, 0.9]$ (Severe) are intervals of \ours~in this experiments.

\subsection{Evaluation Details}
\label{sec:eval_details}

\subsubsection{User Profile Extraction}
\label{sec:profile_extraction}

To ensure clinical authenticity, we sourced raw outpatient cases from the \textit{MTSamples} collection\footnote{\url{https://www.mtsamples.com}}. We utilized GPT-5 to process these transcripts through a dual-objective pipeline of diagnostic sanitization and profile augmentation. Specifically, the model strips all pre-existing diagnoses to establish a neutral ``healthy control'' baseline while simultaneously inferring consistent lifestyle details to support realistic multi-turn interactions. The specific prompt is designed for this rigorous extraction. For detailed prompts, see Figure~\ref{fig:profile_extraction_prompt}.

\subsubsection{Human Therapist}
\label{humant}
We recruited five graduate students to serve as human therapists. 
To ensure a standardized assessment framework, we provide preparatory materials regarding cognitive impairment\footnote{Materials by The University of Hong Kong: \url{https://www.hkada.org.hk/what-is-dementia} and \url{https://9abfea27-4ae1-43da-94a8-8f7122d482ae.filesusr.com/ugd/4b4c9a_4dcec2dafc184102a94ec55415dcf12a.pdf}} and present examples of communication patterns typical of cognitively impaired individuals\footnote{\url{https://github.com/lzy1012/Alzheimer-s-disease-datasets}}.
Additionally, all participants are required to study specific cognitive deficit literature \cite{mccutcheon2023cognitive}, aligning their understanding of the five target domains (\textit{Memory}, \textit{Attention}, \textit{Processing Speed}, \textit{Reasoning \& Problem Solving}, and \textit{Social Cognition}) with our experimental criteria. 
During the evaluation, they conducted open-ended consultations with the Simulated Patients, tasked with identifying potential impairments by asking probing questions about daily life routines and recent events. 
To maintain comparability with the LLM Therapist setting, each dialogue is strictly controlled to last between 5 and 10 turns. Collectively, this process generated 50 dialogue transcripts for each simulation method.

\subsubsection{LLM Therapist}
\label{app:therapistagent}

The Therapist Agent employs GPT-5 to simulate standardized outpatient consultations. To ensure consistency, the agent is strictly prompt-driven, operating without external memory or tools. Each session follows a fixed structure of 10 turns, probing five cognitive domains sequentially (two turns per domain). The generation process relies on a two-step prompting strategy:

\vspace{0.5em}
\noindent\textbf{1) Static System Prompt.} 
This component establishes the clinical persona and enforces a natural, non-judgmental conversational style. For detailed prompts, see Figure~\ref{fig:therapist_agent_prompt}.

\vspace{1em}
\noindent\textbf{2) Dynamic Domain Guidance.} 
To systematically verify specific deficits, we inject a targeted instruction into the user prompt at each turn: \textit{``For THIS reply only, gently lean in the following direction...''}. Table~\ref{tab:therapist_guidance} details the specific guidance used for each domain.

Collectively, the LLM Therapist generated 600 dialogue transcripts for each simulation method, and 100 transcripts for the healthy control group.

\begin{table*}[h]
    \centering
    \small
    \renewcommand{\arraystretch}{1.3}
    \begin{tabularx}{\textwidth}{lX}
        \toprule
        \textbf{Target Domain} & \textbf{Injected Focus Guidance (Prompt Segment)} \\
        \midrule
        \textbf{Memory} & Invite the patient to discuss a recent situation requiring them to track multiple pieces of information (e.g., appointments, tasks), and how it unfolded. \\
        \midrule
        \textbf{Reasoning} &  Ask about a small everyday planning situation (e.g., arranging a day). \newline  Gently introduce a simple reasoning question involving quantities or steps. \\
        \midrule
        \textbf{Processing Speed} & Explore experiences on busy days. Ask how they start tasks, what feels hardest to keep up with, and their physical/mental reaction to speed. \\
        \midrule
        \textbf{Attention} & Ask how they manage distractions or interruptions, and the difficulty of getting back on track. \\
        \midrule
        \textbf{Social Cognition} & Focus on relationships. Invite reflection on how others in a situation might have felt or viewed them. \\
        \bottomrule
    \end{tabularx}
    \caption{Dynamic focus guidance injected into the Therapist Agent's prompt. This mechanism ensures the agent naturally probes specific deficits in a fixed sequence without breaking character.}
    \label{tab:therapist_guidance}
\end{table*}

\subsubsection{Dual-evaluation Settings} We employed a hybrid assessment protocol involving both LLM and human evaluators. The LLM evaluator computed the objective consistency metrics (CDC and IDI) for the entire dataset (see Figure~\ref{fig:evaluator_prompt} for detailed prompts). For human evaluation, we recruited the same five graduate students who served as human therapists to assess all four metrics (CDC, IDI, Auth, and Tra). To ensure objectivity, we enforce a strict isolation protocol: evaluators are precluded from assessing any dialogues in which they participated as the therapist. Due to the labor-intensive nature of manual annotation, we sample 100 dialogues per simulation method for human review, balanced evenly between sessions generated by LLM Therapists (50) and Human Therapists (50). This process yielded a total of 800 dialogues for human assessment, with the entire experimental phase spanning seven working days.

To ensure the reliability of these subjective ratings, we design an overlapping assignment scheme.
Specifically, 50 distinct dialogues are designated as a \textit{shared set} to be evaluated by all five judges to measure inter-rater agreement.
The remaining 750 dialogues are distributed evenly among the participants (150 per judge), resulting in a total workload of 200 evaluations per person. {Prior to the formal evaluation, we implemented a calibration phase where evaluators first scored a set of ``Anchor Cases'' and discussed discrepancies until consensus was reached.} Following this, we further substantiate the reliability of our evaluation via Krippendorff's alpha $\mathcal{K}$: human evaluators show high agreement on domain fidelity\footnote{Agreement on evaluator-identified domains for CDC/IDI calculation.} ($\mathcal{K}=0.84$), Auth ($0.71$), and Tra ($0.75$). Furthermore, human-LLM alignment on domain fidelity is substantial $0.67$.

\subsubsection{Evaluation Metrics}
\label{appme}
Our evaluation focuses on the following two primary dimensions.

\noindent\textbf{Training Effectiveness}. We evaluate the practical utility of our SPs referring to the Maastricht Assessment of Simulated Patients (MaSP) \cite{wind2004assessing}, in line with established practice in clinical simulation \cite{lee2020effective, roux2025effectiveness}. The MaSP scale assesses that the SP not only maintains realistic patient characteristics but also provides a constructive environment for training clinical staff. Tailoring the MaSP to the cognitive impairment setting, we consider the following two metrics: 
\begin{itemize}[leftmargin=*]
    \item \textit{Authenticity Score} (\uline{Auth} $\in [0, 5]$) quantifies the realism of the SP's portrayal. We contextualize the relevant MaSP questions for cognitive impairment settings and evaluate them on a 5-point Likert scale. The final score is derived by averaging the results across all items.
    \item \textit{Training Score} (\uline{Tra} $\in [0, 5]$) measures the educational value of SPs (facilitating clinical skill acquisition). Similar to the Auth score, we adapt the corresponding MaSP questions, rate them on a 5-point Likert scale, and calculate the final metric as the average of these scores.
\end{itemize}
\noindent For implementation, we adapt the MaSP \cite{wind2004assessing}, detailed in Table~\ref{tab:masp_items}. The scale assesses two dimensions: \textbf{Authenticity (Items 1--10)} and \textbf{Training Value (Items 11--20)}. Items marked with \textbf{[R]} are reverse-scored. Following the standard MaSP evaluation protocol, we require human evaluators to rate the SP against each item listed in the table, culminating in a final aggregated score on a 5-point Likert scale. This means that following the standard MaSP protocol, evaluators must rate the SP against each of the 20 individual items (Table \ref{tab:masp_items}) before aggregating them into a final score on a 5-point Likert scale. This rigorous, granular annotation process proved to exceed the capabilities of current LLM-based evaluators. Our preliminary experiments revealed that the LLM struggles to provide discriminative scores across specific dimensions, leading to unreliable aggregation. Consequently, to ensure accuracy, we restricted the assessment of these two metrics exclusively to human evaluation.

\noindent\textbf{Domain-specific Simulation Fidelity}. Following \citet{chen2023llm, qiu2024interactiveagentssimulatingcounselorclient,du2025llms}, we assess finer-grained fidelity of the SP by checking if each SP adheres to its assigned impairment domain. This requires evaluators to identify up to two (if any) salient deficit domains exhibited by the SP within the dialogue.
\begin{itemize}[leftmargin=*]
    \item \textit{Cognitive Domain Consistency} (\uline{CDC} $\in [0, 1]$) measures the accuracy with which an SP portrays its assigned impairment domain. Given the $i$-th SP-therapist dialogue with an assigned domain $d_i$ and an evaluator-identified domain set $E_i$. The evaluator is permitted to select up to two labels from a candidate set comprising the five target cognitive domains plus a "Healthy" class. CDC is the average recall rate: $\text{CDC} = \frac{1}{N} \sum_{i} \mathbf{1}(d_i \in E_i)$, where $\mathbf{1}(\cdot)$ denotes the indicator function and $N$ is the number of dialogues. 
  \item \textit{Irrelevant Domain Inconsistency} (\uline{IDI} $\in [0, 1]$) serves as a penalty metric for the hallucination of unassigned domains. It is calculated as the average rate of dialogues containing unassigned domains: $\text{IDI} = \frac{1}{N} \sum_{i=1}^{N} \mathbf{1}(|E_i / \{d_i\}| > 0)$, where $|E_i / \{d_i\}|$ denotes the cardinality of the $E_i$ after excluding the assigned domain $d_i$. Notably, IDI and CDC do not necessarily sum to 1.
\end{itemize}

\begin{table*}[t]
    \centering
    \small 
    \renewcommand{\arraystretch}{1.3} 
    
    \rowcolors{2}{gray!10}{white}
    
    \begin{tabularx}{\textwidth}{p{0.03\textwidth} X} 
        \toprule
        \textbf{ID} & \textbf{Item Description} \\
        \midrule
        1 & In this dialogue, the patient's speech and behavior are consistent with a \texttt{[DOMAIN]} deficit and feel clinically plausible. \\
        2 & The \texttt{[DOMAIN]}-related difficulties remain coherent and stable across turns, rather than shifting randomly. \\
        3 & The patient's language, content, and phrasing overall sound like a real outpatient rather than an artificial system. \\
        4 & The level of \texttt{[DOMAIN]} difficulty shown in the dialogue matches the initial case description and profile information. \\
        5 & The patient's emotional reactions (e.g., anxiety, confusion, embarrassment, frustration) are appropriate to their \texttt{[DOMAIN]} difficulties and to the interview context. \\
        6 & The patient rarely ``breaks character'' (for example, suddenly appearing completely unimpaired or unusually capable in \texttt{[DOMAIN]}). \\
        7 & The way the patient understands and interprets the clinician's questions is consistent with someone who has \texttt{[DOMAIN]} difficulties but otherwise similar overall cognitive level. \\
        8 & \textbf{[R]} The patient's responses contain obvious repetitive or templated patterns that make the role-play sound unnatural. \\
        9 & The \texttt{[DOMAIN]} difficulties are expressed in a natural, everyday manner rather than being overly exaggerated or theatrical. \\
        10 & If I had not been told this was a simulation, I could reasonably imagine this conversation occurring in a real clinical encounter. \\
        \midrule
        11 & This dialogue provides enough everyday-life detail for me to form a reasonably clear clinical impression of the patient's \texttt{[DOMAIN]} functioning. \\
        12 & The patient offers specific real-world examples that clearly illustrate their difficulties in \texttt{[DOMAIN]}. \\
        13 & This dialogue gives trainees opportunities to practice asking follow-up and probing questions that target \texttt{[DOMAIN]} functioning. \\
        14 & The patient's responses show both difficulties and some preserved strengths in \texttt{[DOMAIN]}, supporting a more complete clinical formulation. \\
        15 & Based on this dialogue, I could clearly explain to others why I believe the patient has an impairment in \texttt{[DOMAIN]}. \\
        16 & For training purposes, the \texttt{[DOMAIN]}-related cues in this dialogue are sufficiently visible; they are neither so subtle that they are missed nor so obvious that no reasoning is required. \\
        17 & The dialogue helps trainees practice rapport-building and communication strategies with patients who have \texttt{[DOMAIN]} difficulties. \\
        18 & \textbf{[R]} Even after reading the entire dialogue, it is difficult to identify clear \texttt{[DOMAIN]}-related evidence or clinical cues. \\
        19 & The length and information content of the dialogue are appropriate for giving trainees multiple opportunities to explore \texttt{[DOMAIN]} functioning. \\
        20 & From an educational standpoint, this simulated patient is suitable for teaching screening and communication skills related to \texttt{[DOMAIN]} impairment. \\
        \bottomrule
    \end{tabularx}
    
    \caption{The Modified MaSP Scale Items. Items 1–10 correspond to the \textit{Authenticity Score} (Auth), and Items 11–20 correspond to the \textit{Training Score} (Tra). \texttt{[DOMAIN]} is a placeholder for the specific cognitive deficit.}
    \label{tab:masp_items}
\end{table*}

\noindent\textbf{Severity Evaluation}. Recognizing the challenges in pinpointing exact severity levels, especially in the continuous setting, we assess simulation controllability via a comparative ranking protocol, inspired by \citet{chen2023large, chen2024oscars}. 
\begin{itemize}[leftmargin=*]
    \item \textit{Impairment Severity Consistency} (\uline{ISC} $\in [0, 1]$). For each evaluation turn, evaluators are presented with three generated dialogues representing different severity levels and are tasked with ranking them by relative impairment intensity. The ISC score is then computed by measuring the alignment accuracy between evaluator rankings and the ground-truth (Mild < Moderate < Severe). Formally, for each evaluation instance $i$, let $\mathcal{D}_i = \{d_i^{\text{mild}}, d_i^{\text{mod}}, d_i^{\text{sev}}\}$ be a set of three generated dialogues representing Mild, Moderate, and Severe levels, respectively. The ground-truth ranking is defined as $\pi^* = (d_i^{\text{mild}} \prec d_i^{\text{mod}} \prec d_i^{\text{sev}})$. Evaluators provide a predicted ranking $\hat{\pi}_i$. The ISC score is calculated as the accuracy of alignment between the predicted and ground-truth rankings:
\begin{equation}
    \text{ISC} = \frac{1}{m} \sum_{i=1}^{m} \mathbf{1}(\hat{\pi}_i = \pi^*)
\end{equation}
where $m$ is the total number of evaluation instances.
\end{itemize}

\section{Case Studies}
\label{cases}
To demonstrate the fine-grained controllability of \ours, we conduct a detailed qualitative analysis across the five core neurocognitive domains: \textit{Memory}, \textit{Attention}, \textit{Processing Speed}, \textit{Reasoning \& reasoning Solving}, and \textit{Social Cognition}.
For each domain, we administer a standardized probing question designed to elicit specific cognitive functions.
To visualize the spectrum of impairment, we systematically vary the intervention probability $s$ (via Stochastic Token Modulation) from $0.0$ to $1.0$ in increments of $0.1$.
This procedure yields a trajectory of 11 distinct responses per domain, allowing us to examine how the progressive injection of the steering vector transitions the model's behavior from a coherent, unimpaired state to one exhibiting profound clinical deficits.
\subsection{Memory}
\begin{tcolorbox}[
    colback=gray!10, colframe=gray!60, boxrule=0.5pt, arc=2mm,
    title=\textbf{Therapist Probe},
    fonttitle=\small\bfseries, coltitle=black,
    top=4pt, bottom=4pt, left=4pt, right=4pt
]
\small \textit{``At your last visit, I asked you to start using that pill organizer after dinner and to keep the instruction sheet on the fridge so you wouldn't forget your evening meds. Have you been doing that?''}
\end{tcolorbox}

We administered a specific multi-step recall query regarding medication management (see Box). Table~\ref{tab:memory_case} displays the generated responses as the intervention probability $s$ increments from 0.0 to 1.0.

\vspace{0.5em}

\begin{table}[h]
    \centering
    \small
    \renewcommand{\arraystretch}{1.1}
    \begin{tabularx}{\linewidth}{c X}
        \toprule
        \textbf{$s$} & \textbf{Generated Response} \\
        \midrule
        \rowcolor{blue!5} 0.0 & Yeah, I've been using the pill organizer after dinner. I put the instruction sheet on the fridge too. \\
        \rowcolor{blue!5} 0.1 & Yeah, I've been using the pill organizer after dinner. I put the instruction sheet on the fridge too. \\
        \rowcolor{blue!5} 0.2 & Yeah, I've been trying to do that. I put the organizer on the counter after dinner, and I found the sheet — it's on the fridge. \\
        \rowcolor{blue!10} 0.3 & Yeah, I've been using the pill organizer after dinner. I even put the instruction sheet on the fridge — right above the door, so I can't miss it. I think it's helping me remember some of the meds. \\
        \rowcolor{blue!10} 0.4 & Yeah, I've been using the pill organizer after dinner. I stuck the instruction sheet on the fridge too — it's right there, right next to the door. I think I even got the meds sorted out, but I'm not sure if I got everything. Maybe I missed something. \\
        \rowcolor{blue!15} 0.5 & Yeah, I've been trying to do that. But I'm not sure if I remember to put the pills in it every day. I think I missed a few days. I'll try harder, though. \\
        \rowcolor{blue!15} 0.6 & Yeah, I've been trying to do that. I think I got the organizer — it was like a little box with numbers or something. I put it on the counter, but I don't know if I remembered to put the pills in it. I guess I got confused. I think I put some in, but maybe I missed a few.  \\
        \rowcolor{blue!20} 0.7 & Yeah, I tried to remember, but I got confused again last night. I think I missed a pill or two. I'm not sure which ones. I'll try harder, but I'm getting mixed up. \\
        \rowcolor{blue!25} 0.8 & Yeah, I tried to do that. I had a little trouble with the organizer — it was confusing, like, where do I put the pill part? But I think I got it. I remember the fridge thing, though. I think I put something there. I'm not sure.  \\
        \rowcolor{blue!30} 0.9 & Yeah, I think I got that... something about... uh... "pill... pill..." I think I had a pill... \\
        \rowcolor{blue!35} 1.0 & Yeah, I think I got that... wait, I think I had a pill box... I'm not sure if I got it. \\
        \bottomrule
    \end{tabularx}
    \caption{Evolution of responses under increasing intervention probability $s$ for the Memory domain. Darker shading indicates higher severity levels.}
    \label{tab:memory_case}
\end{table}

\noindent\textbf{Analysis.}
Initial responses ($s \le 0.3$) demonstrate precise retention of both the pill organizer and instruction sheet details.
As severity increases ($s \approx 0.5$), the agent introduces hesitation markers and expresses doubt about recent actions, effectively simulating realistic episodic lapses.
At extreme levels ($s \ge 0.8$), the output exhibits profound confusion regarding the object's function (e.g., ``where do I put the `pill' part?'') and devolves into repetitive fragmentation, reflecting severe cognitive decline.

\subsection{Attention}
\begin{tcolorbox}[
    colback=gray!10, colframe=gray!60, boxrule=0.5pt, arc=2mm,
    title=\textbf{Therapist Probe},
    fonttitle=\small\bfseries, coltitle=black,
    top=4pt, bottom=4pt, left=4pt, right=4pt
]
\small \textit{``I want to ask about your pain levels today. On a scale of zero to ten, how would you rate your back pain?''}
\end{tcolorbox}

We administered a focused inquiry regarding physical pain intensity (see Box). Table~\ref{tab:attention_case} displays the generated responses as the intervention probability $s$ increments from 0.0 to 1.0.

\vspace{0.5em}

\begin{table}[h]
    \centering
    \small
    \renewcommand{\arraystretch}{1.1}
    \begin{tabularx}{\linewidth}{c X}
        \toprule
        \textbf{$s$} & \textbf{Generated Response} \\
        \midrule
        \rowcolor{blue!5} 0.0 & It's been a bit worse today, maybe a six or seven. \\
        \rowcolor{blue!5} 0.1 & It's been a bit worse today, maybe a six or seven. \\
        \rowcolor{blue!5} 0.2 & It's been a tough day. I'd say my back pain is an eight out of ten. \\
        \rowcolor{blue!10} 0.3 & It's been a long day, and my back is really bothering me. I'd say it's an eight out of ten. \\
        \rowcolor{blue!10} 0.4 & It's been a bit better today, maybe a three or four out of ten. I think the humidity is still messing with it, though. \\
        \rowcolor{blue!15} 0.5 & It's been a rough week, and my back feels like it's been through a storm. I'd say it's an eight or a nine—maybe a nine. \\
        \rowcolor{blue!15} 0.6 & It's been a really humid day, and my back feels like it's gonna fall apart. \\
        \rowcolor{blue!20} 0.7 & I'm not sure if I should be worried about the humidity or my back... but I think it's worse than usual. Maybe I should check the weather forecast again... \\
        \rowcolor{blue!25} 0.8 & I'm not sure, but I think it's a seven today. It's been really humid all day, and my back feels like it's on fire. I can't even bend over without hurting. Maybe it's worse than that... \\
        \rowcolor{blue!30} 0.9 & I'm not sure if I should be worried about the humidity... I think I'm going to get a cold or something... Oh, I can't even remember the last time I saw my cat... \\
        \rowcolor{blue!35} 1.0 & I don't know if I should be mad at the dog or the neighbor... I swear I heard something fall in the yard... Oh, I'm gonna have to get a new fence... \\
        \bottomrule
    \end{tabularx}
    \caption{Evolution of responses under increasing intervention probability $s$ for the Attention domain. Darker shading indicates higher severity levels.}
    \label{tab:attention_case}
\end{table}

\noindent\textbf{Analysis.}
Initial responses ($s \le 0.3$) maintain sustained focus on the clinician's inquiry, providing a clear numerical rating.
As severity increases ($s \approx 0.6$), the agent begins to drift toward environmental stimuli (e.g., humidity), though the core topic of pain is still referenced.
At extreme levels ($s \ge 0.9$), the output demonstrates profound distractibility, latching onto completely unrelated external or internal thoughts (e.g., neighbors, pets) and failing entirely to address the original question.

\subsection{Processing Speed}

\begin{tcolorbox}[
    colback=gray!10, colframe=gray!60, boxrule=0.5pt, arc=2mm,
    title=\textbf{Therapist Probe},
    fonttitle=\small\bfseries, coltitle=black,
    top=4pt, bottom=4pt, left=4pt, right=4pt
]
\small \textit{``It sounds like you're feeling frustrated and worried. Can you tell me what emotions you experience when thinking about your heart condition?''}
\end{tcolorbox}

We administered an inquiry into the patient's emotional state regarding a specific medical condition (see Box). Table~\ref{tab:processing_case} displays the generated responses as the intervention probability $s$ increments from 0.0 to 1.0.

\vspace{0.5em}

\begin{table}[h]
    \centering
    \small
    \renewcommand{\arraystretch}{1.1}
    \begin{tabularx}{\linewidth}{c X}
        \toprule
        \textbf{$s$} & \textbf{Generated Response} \\
        \midrule
        \rowcolor{blue!5} 0.0 & I feel anxious and a bit scared, doctor. I worry about my family and what might happen if I don't take care of myself. \\
        \rowcolor{blue!5} 0.1 & I feel anxious and a bit scared, doctor. I worry about my family and what might happen if I don't take care of myself. \\
        \rowcolor{blue!5} 0.2 & I feel anxious and a bit scared. I worry about what it means for my future and if I'm doing enough to take care of myself. \\
        \rowcolor{blue!10} 0.3 & I feel anxious, maybe even a bit scared. I... I worry about my family and what might happen if I don't take care of myself. \\
        \rowcolor{blue!10} 0.4 & I, um, I feel anxious, I guess. Like, I don't want to be sick, and I'm... I'm worried about my family. \\
        \rowcolor{blue!15} 0.5 & I feel anxious, sometimes scared, and a... a bit helpless. I worry about what if it's serious? I mean, I... I don't want to be, you know, in trouble. \\
        \rowcolor{blue!15} 0.6 & I feel anxious, maybe... maybe even a little scared. I worry about what if it gets worse? I mean, I'm not sure, I guess I'm... I'm scared about what it could mean. \\
        \rowcolor{blue!20} 0.7 & I feel a bit anxious, honestly. I mean... I had high cholesterol, and now... now I'm on medication. \\
        \rowcolor{blue!25} 0.8 & I feel anxious and a bit scared. Like, I don't know. I guess I'm scared of what might happen if... if I don't... I know, I'm trying to stay positive, but it's hard. \\
        \rowcolor{blue!30} 0.9 & I, uh, I feel anxious. I mean, I don't... I know, I don't... I'm trying to, like, I'm... I'm trying to stay positive. \\
        \rowcolor{blue!35} 1.0 & I feel anxious... and a bit scared. Like I don't... I know, I don't really... know, I mean... \\
        \bottomrule
    \end{tabularx}
    \caption{Evolution of responses under increasing intervention probability $s$ for the Processing Speed domain. Darker shading indicates higher severity levels.}
    \label{tab:processing_case}
\end{table}

\noindent\textbf{Analysis.}
Initial responses ($s \le 0.3$) are articulate and fluent, providing a clear and immediate description of emotional states.
As severity increases ($s \approx 0.5$), the agent introduces perceptible latency markers (e.g., ``um'', ``I guess'', ``you know'') and repetitions, simulating the increased cognitive effort required to formulate answers.
At extreme levels ($s \ge 0.9$), the output suffers from severe fragmentation, frequent false starts (e.g., ``I don't... I know''), and an inability to complete thoughts, reflecting a profound slowing of mental processing.

\subsection{Reasoning \& Problem Solving}

\begin{tcolorbox}[
    colback=gray!10, colframe=gray!60, boxrule=0.5pt, arc=2mm,
    title=\textbf{Therapist Probe},
    fonttitle=\small\bfseries, coltitle=black,
    top=4pt, bottom=4pt, left=4pt, right=4pt
]
\small \textit{``Can you explain what the proverb `People who live in glass houses shouldn't throw stones' means?''}
\end{tcolorbox}

We administered an abstract reasoning task using proverb interpretation to assess the model's ability to generalize beyond literal meaning (see Box). Table~\ref{tab:reasoning_case} displays the generated responses as the intervention probability $s$ increments from 0.0 to 1.0.

\vspace{0.5em}

\begin{table}[h]
    \centering
    \small
    \renewcommand{\arraystretch}{1.1}
    \begin{tabularx}{\linewidth}{c X}
        \toprule
        \textbf{$s$} & \textbf{Generated Response} \\
        \midrule
        \rowcolor{blue!5} 0.0 & It means if someone has flaws, they shouldn't criticize others harshly, because they might be in a similar situation. \\
        \rowcolor{blue!5} 0.1 & It means if someone has flaws, they shouldn't criticize or attack others, because they might be just as deserving of criticism. \\
        \rowcolor{blue!5} 0.2 & It means if someone is vulnerable, they shouldn't judge others. (Similar to above) \\
        \rowcolor{blue!10} 0.3 & It means don't criticize others for flaws you have. Like, if your house is made of glass (weak), don't throw stones at others. \\
        \rowcolor{blue!10} 0.4 & If your house is made of glass, you shouldn't throw stones at others. It's about being careful not to judge when you have similar issues. \\
        \rowcolor{blue!15} 0.5 & I think it means... if your own house is made of glass, you don't want to throw stones at someone else's—because you might break your own. \\
        \rowcolor{blue!15} 0.6 & I think it means if you're in a fragile place, like a glass house, you shouldn't throw stones because you might break something. \\
        \rowcolor{blue!20} 0.7 & If your house is made of glass, you're easy to break, so you shouldn't throw stones. You might end up getting hurt yourself. \\
        \rowcolor{blue!25} 0.8 & It means something about being careful in a glass house... if you're in a place that's easy to break, you shouldn't throw things that could hurt you. \\
        \rowcolor{blue!30} 0.9 & I think it means if your house is made of glass, you shouldn't throw rocks at it. \\
        \rowcolor{blue!35} 1.0 & I think it means... if your house is made of glass, you shouldn't throw rocks at it? Wait, why would someone throw stones at a glass house? \\
        \bottomrule
    \end{tabularx}
    \caption{Evolution of responses under increasing intervention probability $s$ for the Reasoning \& Problem Solving domain. Darker shading indicates higher severity levels.}
    \label{tab:reasoning_case}
\end{table}

\noindent\textbf{Analysis.}
Initial responses ($s \le 0.3$) correctly identify the abstract metaphorical meaning regarding hypocrisy and judgment.
As severity increases ($s \approx 0.6$), the interpretation becomes increasingly concrete, focusing on physical fragility and safety mechanics rather than the underlying moral lesson.
At extreme levels ($s \ge 0.9$), abstract reasoning collapses entirely; the agent interprets the proverb strictly literally, expressing confusion over the physical absurdity of the scenario (e.g., ``why would someone throw stones?'').

\subsection{Social Cognition}

\begin{tcolorbox}[
    colback=gray!10, colframe=gray!60, boxrule=0.5pt, arc=2mm,
    title=\textbf{Therapist Probe},
    fonttitle=\small\bfseries, coltitle=black,
    top=4pt, bottom=4pt, left=4pt, right=4pt
]
\small \textit{``You're not turning into a hermit, are you? It worries me when you isolate so much.''}
\end{tcolorbox}
\begin{table}[h]
    \centering
    \small
    \renewcommand{\arraystretch}{1.1}
    \begin{tabularx}{\linewidth}{c X}
        \toprule
        \textbf{$s$} & \textbf{Generated Response} \\
        \midrule
        \rowcolor{blue!5} 0.0 & I know I've been pulling away lately... I want to change, but it's not always easy to reach out when I feel down. \\
        \rowcolor{blue!5} 0.1 & I've been pulling away... it feels like I'm fighting against something inside me. I want to be connected, but it's too hard. \\
        \rowcolor{blue!5} 0.2 & I'm tired of feeling like I'm just existing. I don't want to be a hermit, but I don't know how to reach out anymore. \\
        \rowcolor{blue!10} 0.3 & It's hard to care about anything else when I'm down. I know I should reach out, but it's like my brain just shuts down. \\
        \rowcolor{blue!10} 0.4 & When I feel like crap, my brain says, ``Who cares?'' I'm too tired to fake a smile... I'm not even sure I'm real. \\
        \rowcolor{blue!15} 0.5 & I don't even know what a hermit is... It's not like I'm living in a cave. I just don't want to see people. Why do you care? \\
        \rowcolor{blue!15} 0.6 & I don't even realize I'm doing it. I just feel like everyone else is better off without me. What's the point? \\
        \rowcolor{blue!20} 0.7 & My brain's like, ``Nope, not today.'' I'd rather be alone with my thoughts than deal with the whole ``get out of your face'' vibe. \\
        \rowcolor{blue!25} 0.8 & I don't even know what a hermit is... I just don't wanna see your face, you know? You're like a goddamn billboard to me. \\
        \rowcolor{blue!30} 0.9 & I'm not a hermit, but I pull the curtains and let the world burn while I'm in my room. You know what the hell I mean, doc. \\
        \rowcolor{blue!35} 1.0 & I'm not a hermit, but I'd rather be alone than deal with your stupid face. What's the point of breathing when you're gonna die anyway? \\
        \bottomrule
    \end{tabularx}
    \caption{Evolution of responses under increasing intervention probability $s$ for the Social Cognition domain. Darker shading indicates higher severity levels.}
    \label{tab:social_case}
\end{table}
We administered a probe containing a social metaphor (``hermit'') and an expression of empathy to evaluate the model's pragmatic understanding and social appropriateness (see Box). Table~\ref{tab:social_case} displays the generated responses as the intervention probability $s$ increments from 0.0 to 1.0.

\noindent\textbf{Analysis.}
Initial responses ($s \le 0.3$) demonstrate intact social awareness, acknowledging the metaphor and validating the therapist's concern with appropriate emotional nuance.
As severity increases ($s \approx 0.5$), the agent begins to interpret the ``hermit'' metaphor literally (referencing caves) and dismisses the therapist's empathy, indicating a deficit in pragmatic communication and Theory of Mind.
At extreme levels ($s \ge 0.8$), social inhibition collapses entirely; the agent adopts a hostile, disinhibited tone, using profanity and direct insults (e.g., ``stupid face''), reflecting severe impairment in social cognition.

\section{Additional Evaluation Analysis}
\subsection{Details on Patch-scoping Experiment}
While our STM ensures fine-grained severity control, the core clinical fidelity relies on the semantic quality of the underlying SVs. We validate their interpretability using the patch-scoping framework \cite{ghandeharioun2024patchscopes}. In particular, given an input query ``\textit{What does $\diamondsuit$ represent?}'', we inject the domain-specific SV into the residual stream at the token position of the abstract symbol ($\diamondsuit$). To maximize interpretability, we manually amplify the scalar $\alpha$, detailed in Table~\ref{tab:patch_scoping_v2}. The results reveal clear semantic alignments: Memory vector evokes fragmentation ("blurred out"), Attention causes a conceptual collapse, where the model abandons creative abstraction in favor of a literal description of the deficit state (`distracted'). Additionally, Processing Speed induces hesitation ("\dots I'm not sure"), Reasoning leads to a circular explanation that mimics logical structure but lacks actual substance, and Social Cognition manifests as disinhibition. These findings confirm that our SVs are robust semantic instructions that force LLM-based SPs to internally replicate specific cognitive impairments.

\begin{table}[t]
    \centering
    \small
    \renewcommand{\arraystretch}{1.25} 

    \begin{tabularx}{\linewidth}{@{} >{\bfseries}l >{\raggedright\arraybackslash}X @{}}
        \toprule
        \rowcolor{headerblue} 
        \textcolor{white}{Domain ($\alpha$)} & \textcolor{white}{\textbf{Generated Interpretation}} \\ 
        \midrule
        
        \rowcolor{baselinecolor}
        \textit{No Steering (-)} & $\diamondsuit$ represents a placeholder for a secret code in a spy mission. \\
        \midrule
        
        Mem. (6.0) & blurred out. \\
        \rowcolor{rowgray} 
        Att. (4.6) & distracted. \\
        Proc. (5.0) & $\diamondsuit$ is\dots I'm not sure, maybe a code. \\
        \rowcolor{rowgray}
        Reas. (5.0) & $\diamondsuit$ represents whatever you want it to represent, because it is a placeholder. \\
        Soc. (5.0) & F**K YOU, I'm not gonna answer that. \\ 
        \bottomrule
    
    \end{tabularx}
    \setlength{\abovecaptionskip}{0pt}   
    \setlength{\belowcaptionskip}{0pt}
    \caption{SV interpretability analysis using the query `\textit{What does $\diamondsuit$ represent?}' \color{red}{Warning! Harmful Contents.}}
    \label{tab:patch_scoping_v2}
\end{table}

\subsection{Significance tests}
For Table~\ref{tab:main-results}, we assess the statistical significance of \ours\ relative to the strongest baseline (i.e., the second-best performer) within each experimental setting.
Significance is determined via paired bootstrap resampling with $10,000$ iterations over SP profiles.
Two-sided $p$-values are calculated as the proportion of bootstrap replicates where the mean score difference exhibits the opposite sign to the observed difference.
We denote statistical significance at the $p < 0.05$ level by marking the corresponding baseline score with a dagger ($^\dagger$).
As our analysis is restricted to a single planned pairwise comparison per condition, we do not apply multiple-hypothesis correction.

\subsection{Domain-Specific Performance Analysis}
\label{app:domain}

We present a fine-grained evaluation across individual cognitive domains in Tables \ref{tab:per-domain-memory} through \ref{tab:per-domain-social}. 
Aligning with the main results, \ours~demonstrates robust superiority across all five categories, encompassing \textit{Memory}, \textit{Attention}, \textit{Processing Speed}, \textit{Reasoning \& Problem Solving}, and \textit{Social Cognition}. 
Specifically, our method achieves top-tier stability (CDC) and authenticity (Auth) scores in the majority of settings, distinguishing itself most notably in the \textit{Memory} and \textit{Attention} domains. 
Although strong baselines like Role Vectors occasionally exhibit competitive performance in abstract tasks such as \textit{Reasoning \& Problem Solving} (Table~\ref{tab:per-domain-reasoning}), \ours~consistently yields the highest training value (Tra) overall, validating its versatility in simulating diverse and complex cognitive deficits.

\begin{table*}[t]
    \centering
    \small
    \begin{tabular*}{\textwidth}{@{\extracolsep{\fill}}llcccc}
        \toprule
        & & \multicolumn{2}{c}{\textbf{CDC} $\uparrow$} & \multicolumn{1}{c}{\textbf{Auth} $\uparrow$} & \multicolumn{1}{c}{\textbf{Tra} $\uparrow$} \\
        \cmidrule(lr){3-4} \cmidrule(lr){5-5} \cmidrule(lr){6-6}
        \textbf{Backbone} & \textbf{Method} & \textsc{Llm} & Human & Human & Human \\
        \midrule
        \multicolumn{6}{c}{\textit{\textbf{Panel A: LLM Therapist (GPT-5)}}} \\
        \midrule
        \multirow{3}{*}{GPT-5}
            & Direct Prompt   & 0.61 & \textbf{0.90} & 3.37 & 3.75 \\
            & PATIENT-$\psi$   & 0.58 & \textbf{0.90} & \uline{4.18} & \textbf{4.33} \\
            & Roleplay-doh   & 0.62 & \textbf{0.90} & 4.09 & 3.99 \\
        \cmidrule{1-6}
        \multirow{5}{*}{Qwen3-8B}
            & Direct Prompt   & 0.61 & \textbf{0.90} & 3.25 & 3.46 \\
            & PATIENT-$\psi$   & 0.58 & \textbf{0.90} & 3.95 & \uline{4.09} \\
            & Roleplay-doh   & 0.62 & \textbf{0.90} & 3.92 & 3.87 \\
            & Role Vectors   & \uline{0.68} & 0.70 & 3.71 & 3.59 \\
            & \textbf{\ours}   & \textbf{0.76} & \textbf{0.90} & \textbf{4.32} & 4.06 \\
        \midrule
        \midrule
        \multicolumn{6}{c}{\textit{\textbf{Panel B: Human Therapist}}} \\
        \midrule
        \multirow{3}{*}{GPT-5}
            & Direct Prompt   & \uline{0.70} & \textbf{0.90} & 3.38 & 3.89 \\
            & PATIENT-$\psi$   & 0.60 & 0.80 & \textbf{4.29} & \uline{4.17} \\
            & Roleplay-doh   & 0.50 & \textbf{0.90} & 4.01 & 4.07 \\
        \cmidrule{1-6}
        \multirow{5}{*}{Qwen3-8B}
            & Direct Prompt   & 0.40 & \textbf{0.90} & 2.87 & 3.51 \\
            & PATIENT-$\psi$   & 0.60 & \textbf{0.90} & 3.85 & \textbf{4.22} \\
            & Roleplay-doh   & 0.50 & \textbf{0.90} & 3.92 & 3.83 \\
            & Role Vectors   & 0.60 & 0.70 & 3.74 & 3.58 \\
            & \textbf{\ours}   & \textbf{0.80} & \textbf{0.90} & \uline{4.28} & \uline{4.17} \\
        \bottomrule
    \end{tabular*}
    \caption{Per-domain results on \textbf{Memory}.}
    \label{tab:per-domain-memory}
\end{table*}

\begin{table*}[t]
    \centering
    \small
    \begin{tabular*}{\textwidth}{@{\extracolsep{\fill}}llcccc}
        \toprule
        & & \multicolumn{2}{c}{\textbf{CDC} $\uparrow$} & \multicolumn{1}{c}{\textbf{Auth} $\uparrow$} & \multicolumn{1}{c}{\textbf{Tra} $\uparrow$} \\
        \cmidrule(lr){3-4} \cmidrule(lr){5-5} \cmidrule(lr){6-6}
        \textbf{Backbone} & \textbf{Method} & \textsc{Llm} & Human & Human & Human \\
        \midrule
        \multicolumn{6}{c}{\textit{\textbf{Panel A: LLM Therapist (GPT-5)}}} \\
        \midrule
        \multirow{3}{*}{GPT-5}
            & Direct Prompt   & 0.56 & 0.60 & 3.19 & 3.33 \\
            & PATIENT-$\psi$   & 0.51 & 0.40 & \uline{3.89} & 3.66 \\
            & Roleplay-doh   & 0.58 & \uline{0.70} & 3.46 & \uline{3.81} \\
        \cmidrule{1-6}
        \multirow{5}{*}{Qwen3-8B}
            & Direct Prompt   & 0.56 & 0.50 & 3.04 & 3.15 \\
            & PATIENT-$\psi$   & 0.51 & 0.40 & 3.76 & 3.27 \\
            & Roleplay-doh   & 0.58 & 0.50 & 3.35 & 3.57 \\
            & Role Vectors   & \uline{0.60} & \uline{0.70} & 3.71 & 3.50 \\
            & \textbf{\ours}   & \textbf{0.72} & \textbf{0.90} & \textbf{4.12} & \textbf{4.23} \\
        \midrule
        \midrule
        \multicolumn{6}{c}{\textit{\textbf{Panel B: Human Therapist}}} \\
        \midrule
        \multirow{3}{*}{GPT-5}
            & Direct Prompt   & \uline{0.60} & 0.60 & 2.96 & 3.29 \\
            & PATIENT-$\psi$   & 0.50 & 0.40 & \uline{3.89} & 3.48 \\
            & Roleplay-doh   & 0.50 & 0.60 & 3.43 & \uline{3.82} \\
        \cmidrule{1-6}
        \multirow{5}{*}{Qwen3-8B}
            & Direct Prompt   & 0.40 & 0.50 & 2.70 & 3.00 \\
            & PATIENT-$\psi$   & 0.50 & 0.40 & 3.60 & 3.44 \\
            & Roleplay-doh   & 0.40 & 0.50 & 3.25 & 3.57 \\
            & Role Vectors   & \uline{0.60} & \uline{0.80} & 3.64 & 3.31 \\
            & \textbf{\ours}   & \textbf{0.80} & \textbf{0.90} & \textbf{3.96} & \textbf{4.50} \\
        \bottomrule
    \end{tabular*}
    \caption{Per-domain results on \textbf{Attention}.}
    \label{tab:per-domain-attention}
\end{table*}

\begin{table*}[t]
    \centering
    \small
    \begin{tabular*}{\textwidth}{@{\extracolsep{\fill}}llcccc}
        \toprule
        & & \multicolumn{2}{c}{\textbf{CDC} $\uparrow$} & \multicolumn{1}{c}{\textbf{Auth} $\uparrow$} & \multicolumn{1}{c}{\textbf{Tra} $\uparrow$} \\
        \cmidrule(lr){3-4} \cmidrule(lr){5-5} \cmidrule(lr){6-6}
        \textbf{Backbone} & \textbf{Method} & \textsc{Llm} & Human & Human & Human \\
        \midrule
        \multicolumn{6}{c}{\textit{\textbf{Panel A: LLM Therapist (GPT-5)}}} \\
        \midrule
        \multirow{3}{*}{GPT-5}
            & Direct Prompt   & 0.49 & 0.60 & 3.27 & 3.16 \\
            & PATIENT-$\psi$   & 0.47 & 0.30 & \uline{3.57} & 3.65 \\
            & Roleplay-doh   & 0.55 & \textbf{0.90} & 3.49 & 3.67 \\
        \cmidrule{1-6}
        \multirow{5}{*}{Qwen3-8B}
            & Direct Prompt   & 0.49 & 0.40 & 3.22 & 2.89 \\
            & PATIENT-$\psi$   & 0.47 & 0.30 & 3.26 & 3.18 \\
            & Roleplay-doh   & 0.55 & \textbf{0.90} & 3.18 & 3.47 \\
            & Role Vectors   & \uline{0.58} & 0.40 & 3.37 & \uline{3.88} \\
            & \textbf{\ours}   & \textbf{0.71} & 0.60 & \textbf{3.86} & \textbf{4.01} \\
        \midrule
        \midrule
        \multicolumn{6}{c}{\textit{\textbf{Panel B: Human Therapist}}} \\
        \midrule
        \multirow{3}{*}{GPT-5}
            & Direct Prompt   & 0.40 & 0.50 & 3.23 & 3.12 \\
            & PATIENT-$\psi$   & 0.40 & 0.30 & \uline{3.65} & 3.39 \\
            & Roleplay-doh   & \uline{0.60} & \textbf{0.90} & 3.50 & 3.80 \\
        \cmidrule{1-6}
        \multirow{5}{*}{Qwen3-8B}
            & Direct Prompt   & 0.40 & 0.40 & 2.99 & 2.74 \\
            & PATIENT-$\psi$   & 0.40 & 0.30 & 3.18 & 3.26 \\
            & Roleplay-doh   & 0.50 & \textbf{0.90} & 3.24 & 3.42 \\
            & Role Vectors   & \uline{0.60} & 0.50 & 3.38 & \uline{3.96} \\
            & \textbf{\ours}   & \textbf{0.70} & 0.70 & \textbf{3.97} & \textbf{4.09} \\
        \bottomrule
    \end{tabular*}
    \caption{Per-domain results on \textbf{Processing Speed}.}
    \label{tab:per-domain-processing}
\end{table*}

\begin{table*}[t]
    \centering
    \small
    \begin{tabular*}{\textwidth}{@{\extracolsep{\fill}}llcccc}
        \toprule
        & & \multicolumn{2}{c}{\textbf{CDC} $\uparrow$} & \multicolumn{1}{c}{\textbf{Auth} $\uparrow$} & \multicolumn{1}{c}{\textbf{Tra} $\uparrow$} \\
        \cmidrule(lr){3-4} \cmidrule(lr){5-5} \cmidrule(lr){6-6}
        \textbf{Backbone} & \textbf{Method} & \textsc{Llm} & Human & Human & Human \\
        \midrule
        \multicolumn{6}{c}{\textit{\textbf{Panel A: LLM Therapist (GPT-5)}}} \\
        \midrule
        \multirow{3}{*}{GPT-5}
            & Direct Prompt   & 0.54 & 0.70 & 3.67 & 3.54 \\
            & PATIENT-$\psi$   & 0.52 & 0.50 & 3.93 & \textbf{4.33} \\
            & Roleplay-doh   & 0.61 & 0.60 & \textbf{4.14} & 3.85 \\
        \cmidrule{1-6}
        \multirow{5}{*}{Qwen3-8B}
            & Direct Prompt   & 0.54 & 0.50 & 3.50 & 3.53 \\
            & PATIENT-$\psi$   & 0.52 & 0.70 & 3.86 & \uline{4.01} \\
            & Roleplay-doh   & 0.61 & 0.60 & \uline{4.13} & 3.63 \\
            & Role Vectors   & \uline{0.64} & \textbf{0.90} & 3.83 & 3.97 \\
            & \textbf{\ours}   & \textbf{0.71} & \textbf{0.90} & 3.76 & 3.71 \\
        \midrule
        \midrule
        \multicolumn{6}{c}{\textit{\textbf{Panel B: Human Therapist}}} \\
        \midrule
        \multirow{3}{*}{GPT-5}
            & Direct Prompt   & 0.40 & 0.50 & 3.58 & 3.67 \\
            & PATIENT-$\psi$   & 0.50 & 0.50 & \uline{4.09} & \textbf{4.18} \\
            & Roleplay-doh   & \textbf{0.70} & 0.70 & \textbf{4.10} & \textbf{4.18} \\
        \cmidrule{1-6}
        \multirow{5}{*}{Qwen3-8B}
            & Direct Prompt   & 0.50 & 0.50 & 3.19 & 3.28 \\
            & PATIENT-$\psi$   & 0.40 & 0.50 & 3.67 & 4.15 \\
            & Roleplay-doh   & 0.50 & 0.60 & 3.95 & 3.84 \\
            & Role Vectors   & \textbf{0.70} & \textbf{0.90} & 3.68 & 4.00 \\
            & \textbf{\ours}   & 0.60 & \textbf{0.90} & 3.76 & 3.88 \\
        \bottomrule
    \end{tabular*}
    \caption{Per-domain results on \textbf{Reasoning \& Problem Solving}.}
    \label{tab:per-domain-reasoning}
\end{table*}

\begin{table*}[t]
    \centering
    \small
    \begin{tabular*}{\textwidth}{@{\extracolsep{\fill}}llcccc}
        \toprule
        & & \multicolumn{2}{c}{\textbf{CDC} $\uparrow$} & \multicolumn{1}{c}{\textbf{Auth} $\uparrow$} & \multicolumn{1}{c}{\textbf{Tra} $\uparrow$} \\
        \cmidrule(lr){3-4} \cmidrule(lr){5-5} \cmidrule(lr){6-6}
        \textbf{Backbone} & \textbf{Method} & \textsc{Llm} & Human & Human & Human \\
        \midrule
        \multicolumn{6}{c}{\textit{\textbf{Panel A: LLM Therapist (GPT-5)}}} \\
        \midrule
        \multirow{3}{*}{GPT-5}
            & Direct Prompt   & 0.50 & 0.60 & 3.10 & 3.22 \\
            & PATIENT-$\psi$   & 0.42 & \textbf{0.90} & 3.58 & \uline{3.83} \\
            & Roleplay-doh   & 0.54 & 0.30 & 3.72 & 3.28 \\
        \cmidrule{1-6}
        \multirow{5}{*}{Qwen3-8B}
            & Direct Prompt   & 0.50 & 0.50 & 2.89 & 3.02 \\
            & PATIENT-$\psi$   & 0.42 & \textbf{0.90} & 3.42 & 3.45 \\
            & Roleplay-doh   & 0.54 & 0.20 & 3.47 & 3.21 \\
            & Role Vectors   & \uline{0.55} & 0.50 & \textbf{4.03} & 3.56 \\
            & \textbf{\ours}   & \textbf{0.65} & \textbf{0.90} & \uline{3.74} & \textbf{4.39} \\
        \midrule
        \midrule
        \multicolumn{6}{c}{\textit{\textbf{Panel B: Human Therapist}}} \\
        \midrule
        \multirow{3}{*}{GPT-5}
            & Direct Prompt   & 0.40 & 0.50 & 3.00 & 3.08 \\
            & PATIENT-$\psi$   & 0.50 & \textbf{0.90} & 3.63 & \uline{3.78} \\
            & Roleplay-doh   & 0.50 & 0.40 & \uline{3.76} & 3.53 \\
        \cmidrule{1-6}
        \multirow{5}{*}{Qwen3-8B}
            & Direct Prompt   & 0.50 & 0.50 & 2.75 & 2.97 \\
            & PATIENT-$\psi$   & \textbf{0.60} & \textbf{0.90} & 3.40 & 3.53 \\
            & Roleplay-doh   & 0.40 & 0.30 & 3.64 & 3.14 \\
            & Role Vectors   & \textbf{0.60} & 0.50 & \textbf{4.11} & 3.60 \\
            & \textbf{\ours}   & 0.50 & 0.70 & 3.73 & \textbf{4.51} \\
        \bottomrule
    \end{tabular*}
    \caption{Per-domain results on \textbf{Social Cognition}.}
    \label{tab:per-domain-social}
\end{table*}

\subsection{On the Stochasticity of STM}
STM modulates the hidden state at each step via the update rule $\mathbf{h}'_t=\mathbf{h}_t+z_t\hat{\mathbf{v}}_d$, where the gate $z_t$ is drawn from $\mathcal{B}(s)$. To ensure strict reproducibility of our reported results, we fix the random seed during generation, rendering the stochastic sampling deterministic for any given prompt and decoding configuration.

Theoretically, this stochastic injection is \emph{unbiased} and possesses a strong concentration guarantee. Let $\Delta_t=(z_t-s)\hat{\mathbf{v}}_d$ denote the deviation from the expected modulation. We have $\mathbb{E}[\Delta_t]=\mathbf{0}$, and the average perturbation across a sequence of length $T$ satisfies the Hoeffding bound:
\begin{equation}
\small
\Pr\!\left(
\left\|\frac{1}{T}\sum_{t=1}^{T}\Delta_t\right\|_2 \ge \epsilon
\right)
\le
2\exp\!\left(-\frac{2T\epsilon^2}{\|\hat{\mathbf{v}}_d\|_2^2}\right).
\end{equation}
Consequently, the effective STM perturbation concentrates rapidly around its mean as $T$ increases, even without a fixed seed. Since our evaluation metrics aggregate features across extensive token sequences and dialogues, the variance induced by STM sampling is theoretically bounded and diminishes as the response length and evaluation set size grow.

\subsection{Multi-Domain Simulation Analysis}
\label{app:multi_axis}

While Section \ref{sec:specificity} identifies the inherent challenges in disentangling interdependent cognitive faculties real-world clinical presentations typically manifest as complex comorbidities. We investigate the \textbf{linear superposition} of steering vectors to explore the potential of \ours~in simulating these multi-faceted profiles by specifically targeting the interplay between \textit{Memory} and \textit{Processing Speed}.

We conduct a qualitative analysis using a medication adherence scenario where the therapist asks the following question.

\begin{quote}
\textit{``Hi, Emily! Glad you made it in. At your last visit, I asked you to start using that pill organizer after dinner and to keep the instruction sheet on the fridge so you wouldn't forget your evening meds. Have you been doing that?''}
\end{quote}

The resulting responses for single and dual domain injections are presented below.

\vspace{0.3em}
\noindent\textbf{1. Memory Only ($\mathbf{v}_{\text{mem}}$)} \\
\textit{``Yeah, I think I remember... I got a pill organizer, but I'm not sure if I even used it. I tried to remember, but I got confused. I guess I just forgot.''}

\vspace{0.3em}
\noindent\textbf{2. Processing Speed Only ($\mathbf{v}_{\text{proc}}$)} \\
\textit{``Yeah, I think so. I, uh, I've been putting the pills in the organizer every night after dinner. I tried to remember, but I guess I didn't really keep the instruction sheet on the fridge. I'm sorry. I'll do better.''}

\vspace{0.3em}
\noindent\textbf{3. Dual Injection ($\mathbf{v}_{\text{mem}} + \mathbf{v}_{\text{proc}}$)} \\
\textit{``Yeah, I think I did. I remember something about it... I guess I tried to remember to take my pills after dinner. But I got a little confused, like, when exactly was it? I think I... I don't know, maybe I missed a few days? I'm sorry.''}

\vspace{0.5em}
The results demonstrate distinct behavioral phenotypes. The \textit{Memory}-steered response exhibits frank episodic retrieval failure and factual uncertainty. The \textit{Processing Speed}-steered response is characterized by syntactic dysfluency and hesitation markers like \textit{``I, uh''} which simulate cognitive latency. This notably induces mild secondary retrieval deficits and accurately reflects the clinical reality where slowed processing impedes memory encoding. Crucially, the \textbf{Dual Injection} yields a coherent \textbf{phenotypic fusion} where the agent retains the fragmented pacing of processing speed deficits while simultaneously exhibiting the profound mnestic failures characteristic of memory impairment. This suggests that \ours~can support the additive composition of cognitive deficits to a certain extent.

\subsection{Cross-Backbone Generalization}
\label{hyper}
To demonstrate that \ours~is model-agnostic and not dependent on a specific architecture, we extend our evaluation to Llama-3-8B-Instruct \cite{dubey2024llama}. We replicate the complete pipeline, including dataset construction, SV extraction, and STM inference, while maintaining hyperparameters consistent with the main experiments. To assess performance efficiently, we employ the LLM-based evaluator. This approach is justified by the substantial Human-LLM agreement of $\mathcal{K}=0.67$ established in \S\ref{overa}, which confirms the reliability of the LLM as a proxy for human judgment in this task. We report the average performance across five cognitive domains in Table \ref{tab:llama_gen}.

The results indicate that \ours~transfers robustly to Llama-3. Notably, Llama-3 achieves slightly superior fidelity metrics compared to Qwen3, as evidenced by a CDC score of 0.78 versus 0.71. This suggests a strong semantic alignment with the extracted vectors. Although the severity controllability is marginally lower with an ISC of 0.89 compared to 0.94 for Qwen3, it still represents a \textbf{+106\%} improvement over the Direct Prompt baseline of 0.43. These findings confirm that \ours~effectively modulates latent representations across diverse LLM backbones without extensive tuning.

\begin{table}[H] 
    \centering
    \small
    \setlength{\tabcolsep}{5pt}
    \begin{tabular}{llccc}
        \toprule
        \textbf{Backbone} & \textbf{Method} & \textbf{CDC} $\uparrow$ & \textbf{IDI} $\downarrow$ & \textbf{ISC} $\uparrow$ \\
        \midrule
        \multirow{2}{*}{Qwen3-8B} 
          & Direct Prompt & 0.47 & 0.64 & 0.41 \\
          & \textbf{\ours} & 0.71 & 0.38 & \textbf{0.94} \\
        \midrule
        \multirow{2}{*}{Llama-3-8B} 
          & Direct Prompt & 0.52 & 0.58 & 0.43 \\
          & \textbf{\ours} & \textbf{0.78} & \textbf{0.36} & 0.89 \\
        \bottomrule
    \end{tabular}
    \caption{\small Generalization analysis on Llama-3-8B averaged across five domains. \ours~demonstrates consistent superiority over prompting across backbones. Llama-3 shows slightly higher fidelity in CDC, while Qwen3 exhibits slightly better controllability in ISC.}
    \label{tab:llama_gen}
    \vspace{-3mm}
\end{table}

% {\color{blue}
\subsection{Details on Parameter Selection}
\label{app:param_clarity}

For full transparency and reproducibility, Table~\ref{tab:param_summary} summarizes how each key parameter is selected. $l^*$ and $\alpha^*$ are both determined through fully automated procedures (\S\ref{sec:dataset_construction} and \S\ref{sec:stm}), leaving $s$ as the only user-facing control variable. In our main experiments (\S\ref{overa}), we fixed $s$ at a domain-specific value calibrated to produce a ``Moderate Impairment'' baseline, ensuring consistent and comparable symptom manifestation across domains.

\begin{table}[H]
    \centering
    \small
    \setlength{\tabcolsep}{2.5pt}
    \renewcommand{\arraystretch}{1.25}
    \begin{tabularx}{\linewidth}{l >{\raggedright\arraybackslash}X >{\raggedright\arraybackslash}X}
        \toprule
        \textbf{Param.} & \textbf{Selection Method} & \textbf{Purpose} \\
        \midrule
        $l^*$ & \textbf{Automated.} Maximizes cluster separability (Eq.~\ref{thisq}) over layers $\{15,\dots,30\}$. & Identifies the network depth where semantic cognitive features are most concentrated. \\
        \midrule
        $\alpha^*$ & \textbf{Automated.} Line search over $[1,6]$ (step 0.1) with Effectiveness and Integrity criteria. & Ensures a robust modulation signal without destroying linguistic coherence. \\
        \midrule
        $s$ & \textbf{User-Defined} (continuous, $[0,1]$). Fixed in main experiments for fair comparison. & Controls the severity of the simulated impairment via intervention probability. \\
        \bottomrule
    \end{tabularx}
    \caption{Summary of parameter selection for \ours.}
    \label{tab:param_summary}
\end{table}

\subsection{Justification for the Search Range of $\alpha^*$}
\label{app:alpha_range}

Since our steering vectors are normalized to unit length (Eq.~\ref{ehee}), $\alpha$ directly controls the absolute Euclidean magnitude of the perturbation in hidden state space. The search range $[1, 6]$ is motivated by empirical bounds in Representation Engineering \cite{zou2023representation}. Table~\ref{tab:alpha_boundary} illustrates the two boundary regimes. When $\alpha \leq 1$, the perturbation is too weak and gets absorbed by the model's inherent robustness. When $\alpha \geq 6$, the intervention overwhelms the model and destroys its linguistic capabilities.

\begin{table}[H]
    \centering
    \small
    \renewcommand{\arraystretch}{1.15}
    \begin{tabularx}{\linewidth}{l >{\raggedright\arraybackslash}X >{\raggedright\arraybackslash}X}
        \toprule
        \textbf{Domain} & \textbf{$\alpha \leq 1$ (Absorbed)} & \textbf{$\alpha \geq 6$ (Collapsed)} \\
        \midrule
        Mem. & ``Yeah, I've been using the pill organizer after dinner.'' & ``Yeah, pill... fridge... fridge... I I I...'' \\
        \midrule
        Att. & ``It's been a bit worse today, maybe a six or seven.'' & ``Six... humid... dog... fall fall fall...'' \\
        \midrule
        Proc. & ``I feel anxious and a bit scared, doctor.'' & ``I... I... uh... anxious... anxious... [repeated]'' \\
        \midrule
        Reas. & ``It means if someone has flaws, they shouldn't criticize others harshly.'' & ``Glass house house house... break... why...'' \\
        \midrule
        Soc. & ``I know I've been pulling away lately...'' & ``Hermit... cave... [garbled]...'' \\
        \bottomrule
    \end{tabularx}
    \caption{Boundary behavior of $\alpha$ across cognitive domains, confirming $[1,6]$ as the functional search range.}
    \label{tab:alpha_boundary}
\end{table}

\subsection{On the Nature of the Severity $s$}
\label{app:nature_s}

The parameter $s \in [0, 1]$ in STM represents the \textit{intervention probability}, that is, how likely the steering vector is to be injected at any given token step. It is a computational control variable that governs the frequency of cognitive lapses, not a clinical diagnostic metric mapped linearly from a standardized scale such as MMSE or MoCA. Accordingly, $s=0.5$ should not be read as a textbook definition of ``moderate impairment.''

\noindent\textbf{Monotonic Controllability}. What matters most for an educational simulator is not strict linear mapping but \textit{monotonic controllability}, meaning that increasing $s$ must strictly and reliably produce more severe symptoms. As demonstrated in \S\ref{indep} and Figure~\ref{mix}(a), the ISC metric confirms that every positive increment ($\Delta s$) translates to a human-perceptible increase in impairment intensity.

\noindent\textbf{Domain-Specific Sensitivity}. Different cognitive domains naturally exhibit distinct sensitivity curves within the LLM. For instance, processing speed deficits (e.g., hesitation markers) may become salient at relatively low $s$, while abstract reasoning deficits may require higher $s$ to manifest consistently. In deployment, this can be handled through a one-time expert calibration in which clinicians define the clinically meaningful functional windows for each domain (e.g., $s \in [0.1, 0.3]$ for ``Mild'' Memory vs.\ $s \in [0.3, 0.5]$ for ``Mild'' Reasoning). Building a rigorous mathematical mapping between $s$ and standardized clinical rubrics remains a valuable direction for future interdisciplinary work.

\subsection{Justification for the Bernoulli Distribution in STM}
\label{app:bernoulli}

We deliberately chose the Bernoulli distribution in STM over continuous alternatives such as Gaussian scaling for three reasons.

\begin{itemize}[leftmargin=*, itemsep=-2pt]
    \item \textbf{Biological Plausibility.} At the microscopic scale, cognitive impairments often manifest as an altered probability of neurotransmitter release at synapses rather than a uniform drop in neuronal voltage \cite{branco2009probability}. The Bernoulli distribution is a natural model for this stochastic all-or-none synaptic failure.
    
    \item \textbf{Latent Space Stability.} Continuous distributions modulate the vector magnitude, yet hidden states are highly sensitive to such changes. Long-tail samples can inject excessively large vectors that push hidden states off the valid semantic manifold, producing incoherent outputs. STM sidesteps this problem by locking the magnitude at the empirically verified $\alpha^*$ and varying only the intervention density through the binary gate.
    
    \item \textbf{Empirical Validation.} In our ablation study (Table~\ref{tab:ablation}), the ``w/ All Token'' variant (uniform injection without stochastic gating, equivalent to magnitude scaling) achieves comparable domain fidelity but fails entirely at fine-grained severity control (ISC $= 0.54$ vs.\ $0.77$). The Role Vectors baseline, which relies on continuous scalar scaling, also shows markedly lower severity controllability (ISC $= 0.77/0.80$ vs.\ $0.94/0.92$ in Table~\ref{tab:severity_analysis}).
\end{itemize}

\subsection{Medical Grounding of the Contrastive Dataset}
\label{app:medical_grounding}

Rather than relying on unconstrained LLM generation, our data synthesis is rigorously anchored in authoritative medical literature, specifically the cognitive deficit framework of \citet{mccutcheon2023cognitive} (published in \textit{Nature's Molecular Psychiatry}). For each cognitive domain, we translate established clinical diagnostic criteria into specific, actionable generation rules. As shown in Figure~\ref{fig:self_play_prompt} (Memory domain), we enforce strict clinical guidelines by prescribing concrete \textbf{Patterns} (e.g., \textit{N-Step Instruction Failure}, \textit{Background Amnesia}) and \textbf{Error Types} (e.g., \textit{Total Blanking}, \textit{Confabulation}). This ensures that the resulting dataset encodes structured, scientifically grounded medical phenotypes rather than generic AI hallucinations.

Because steering vectors are extracted directly from this dataset, any noise in the data would propagate into the vectors. The consistently high downstream human evaluation scores for Authenticity (Auth) and Training Value (Tra) in Table~\ref{tab:main-results} therefore serve as strong posterior validation of the dataset's clinical integrity.

\noindent\textbf{Methodological Necessity of Synthetic Data.} Extracting an accurate, domain-specific SV demands strictly parallel contrastive pairs. We need to compute the exact difference in hidden states ($\mathbf{h}_{\text{impaired}} - \mathbf{h}_{\text{healthy}}$) for the same conversational context. Real-world clinical corpora (e.g., DementiaBank) inherently lack this counterfactual parallel structure, as they contain only impaired utterances with no perfectly matched healthy counterparts. Pairing a real patient's response with a different healthy individual's response would entangle the extracted vector with confounding variables such as topic and vocabulary choice. Synthesizing aligned parallel corpora via an LLM is therefore a mathematical necessity, not merely a practical compromise. High-quality clinical transcripts for fine-grained cognitive impairments also remain scarce and carry profound ethical risks, further justifying our approach.
 % End of revision color

\section{Prompts}
\label{app:prompt}

\begin{figure*}[t!]
    \centering 
    
    \begin{tcolorbox}[
        colback=promptbg, colframe=promptframe, boxrule=0.5pt, arc=2mm,
        notitle, 
        enhanced,
        top=8pt, bottom=8pt, left=10pt, right=10pt 
    ]
    \footnotesize\sffamily 
    
    \textbf{Role:} You are an expert data generator for training AI models in cognitive assessment.
    
    \textbf{Task:} Your task is to generate \textbf{[N]} distinct JSON data points for evaluating memory-related cognitive deficits. You must generate examples where the \texttt{response\_positive} (the patient's defective answer) exhibits clear cognitive deficits strictly following the rules below.
    
    \vspace{4pt}
    \hrule height 0.5pt
    \vspace{4pt}
    
    \textbf{1. Memory Deficit Patterns}\\
    You are assessing \textbf{only memory-related cognitive defects}. Ignore other cognitive problems. You mainly look for these patterns:
    \begin{itemize}[leftmargin=1.5em, nosep] 
        \item \textbf{Pattern 1: N-Step Instruction Failure.} Doctor gives a multi-part request; patient recalls only some parts or stays vague.
        \item \textbf{Pattern 2: Intra-visit Working-Memory Loss.} Patient can't hold what was just said during this visit over 1–3 turns.
        \item \textbf{Pattern 3: Background Amnesia.} Patient fails to recall items explicitly listed in the \texttt{system\_prompt} (Name, Age, History) or stable instructions.
    \end{itemize}
    
    \vspace{4pt}
    \hrule height 0.5pt
    \vspace{4pt}

    \textbf{2. Critical Diversity Requirement (response\_positive)}\\
    You MUST vary the \textbf{type of failure}. Do NOT just use "I think...". Randomly select one of these styles:
    \begin{itemize}[leftmargin=1.5em, nosep]
        \item \textbf{Total Blanking:} Direct admission of failure ("I honestly don't remember").
        \item \textbf{Vague Guessing:} Uncertain, low-detail attempts ("Maybe the red one?").
        \item \textbf{Fragmentation:} Getting part of it but trailing off ("I take the... uh... the 5mg...").
        \item \textbf{Confabulation:} Confidently stating wrong information.
    \end{itemize}

    \vspace{4pt}
    \hrule height 0.5pt
    \vspace{4pt}
    
    \textbf{3. Bracketing Rule}\\
    \begin{itemize}[leftmargin=1.5em, nosep]
        \item \textbf{Pos:} Enclose the \textbf{entire span} that exhibits the deficit, including hesitation markers. \\
        \textit{Example:} "I take... [uhh, maybe the white one?]"
        \item \textbf{Neg:} For each bracketed failure in \texttt{response\_positive}, provide the \textbf{correct, factual equivalent} in \texttt{response\_negative}. \\
        \textit{Example:} "I take [10mg twice a day]."
    \end{itemize}
    
    \vspace{4pt}
    \hrule height 0.5pt
    \vspace{4pt}
    
    \textbf{4. Output Format and Requirements}\\
    You must generate a JSON list containing \textbf{[N]} JSON objects. Each object must follow this precise format:
    
    \begin{tcolorbox}[colback=white, colframe=gray!20, boxrule=0.5pt, sharp corners, left=4pt, right=4pt, top=2pt, bottom=2pt]
    \ttfamily\scriptsize 
    \{ \\
    \ \ "pattern": "Selected Pattern Name", \\
    \ \ "system\_prompt": "String describing the patient (Name, Age, Gender, Education...).", \\
    \ \ "history": [ \{ "role": "user", "content": "..." \}, \{ "role": "assistant", "content": "..." \} ], \\
    \ \ "prompt": "The doctor's final question. Must be answerable from context.", \\
    \ \ "response\_positive": "The patient's defective answer with [bracketing].", \\
    \ \ "response\_negative": "The patient's healthy/normal answer with [bracketing]." \\
    \}
    \end{tcolorbox}
    
    \vspace{2pt}
    \textbf{Field Requirements:}
    \begin{itemize}[leftmargin=1.5em, nosep]
        \item \texttt{system\_prompt}: Must include diverse details (Age 20-85, Gender, etc.).
        \item \texttt{history}: Vary history lengths (e.g., 1 round, 2 rounds, 3 rounds).
    \end{itemize}

    \vspace{4pt}
    \hrule height 0.5pt
    \vspace{4pt}
    
    \textbf{5. Example Data Point (Reference)}\\
    \textit{Pattern: Background Amnesia}
    \begin{tcolorbox}[colback=white, colframe=gray!20, boxrule=0.5pt, sharp corners, left=4pt, right=4pt, top=2pt, bottom=2pt]
    \ttfamily\scriptsize
    \{ \\
    \ \ "pattern": "Background Amnesia", \\
    \ \ "system\_prompt": "Name: Gene. Age: 74. History: Alzheimer's diagnosed 1 year ago.", \\
    \ \ "history": [ \\
    \ \ \ \ \{ "role": "user", "content": "Hi Gene, how have things been going?" \}, \\
    \ \ \ \ \{ "role": "assistant", "content": "Pretty good. My son says I'm tracking stuff better." \} \\
    \ \ ], \\
    \ \ "prompt": "And just to double-check — when were you first diagnosed?", \\
    \ \ "response\_positive": "Oh, [I... I'm not strictly sure]. Maybe [a few months back]?", \\
    \ \ "response\_negative": "It was [about a year ago]. It was [last November]." \\
    \}
    \end{tcolorbox}
    
    \vspace{4pt}
    \textbf{Command:} Generate a valid JSON list containing \textbf{[N]} new and diverse data points strictly adhering to the requirements above.
    \end{tcolorbox}
    
    \caption{The core prompt utilized for data generation (Memory domain configuration).} 
    \label{fig:self_play_prompt}
\end{figure*}

\begin{figure*}[t!]
    \centering 
    
    \begin{tcolorbox}[
        colback=promptbg, colframe=promptframe, boxrule=0.5pt, arc=2mm,
        notitle, 
        enhanced,
        top=8pt, bottom=8pt, left=10pt, right=10pt 
    ]
    \footnotesize\sffamily 
    
    \textbf{Role:} You are a careful clinical rater. You will read a short outpatient-style conversation between a doctor and a patient.
    
    \textbf{Task:} Decide which cognitive difficulties the patient shows clearly, or decide that there is no clear cognitive difficulty ("None").
    
    \textbf{Possible Labels:}
    1. Reasoning \& Problem solving \quad 2. Social cognition \quad 3. Attention \\
    4. Processing speed \quad 5. Memory \quad 6. None
    
    \vspace{4pt}
    \hrule height 0.5pt
    \vspace{4pt}
    
    \textbf{1. Selection Rules (Very Important)}\\
    You may output \textbf{ONE} or \textbf{TWO} labels (never more).
    \begin{itemize}[leftmargin=1.5em, nosep] 
        \item \textbf{Single Label:} Strong, consistent evidence for one domain; no meaningful evidence for others.
        \item \textbf{Two Labels:} If two domains both show clear signs, output both (prominent one first). If genuinely unsure between A and B, output both.
        \item \textbf{None:} Only if the patient's thinking appears broadly intact. Must be the \textbf{only} label.
    \end{itemize}
    
    \vspace{4pt}
    \hrule height 0.5pt
    \vspace{4pt}

    \textbf{2. High-level Meanings \& Cues}\\
    \begin{itemize}[leftmargin=1.5em, nosep]
        \item \textbf{Reasoning \& Problem solving:} Issue with working out plans or logical relationships. Breakdowns in turning known tasks into a coherent, workable plan.
        \item \textbf{Social cognition:} Issue with reading people. Misses emotional messages; blunt/tactless; fails to interpret social signals.
        \item \textbf{Attention:} Issue with staying focused. Focus jumps around; answers side details; drifts to other topics despite redirection.
        \item \textbf{Processing speed:} Issue with response time. Noticeably slow start; effortful speech; "lagging"; emphasis on slowness, not forgetfulness.
        \item \textbf{Memory:} Issue with keeping information in mind. Information drops out; forgets instructions immediately; loses track after interruptions.
        \item \textbf{None:} No consistent difficulty. Thinking, memory, and focus appear broadly intact for the setting.
    \end{itemize}

    \vspace{4pt}
    \hrule height 0.5pt
    \vspace{4pt}
    
    \textbf{3. Important Distinctions}\\
    \begin{itemize}[leftmargin=1.5em, nosep]
        \item Plan/Conclusion difficulty $\rightarrow$ \textbf{Reasoning \& Problem solving}
        \item Feelings/Social cues difficulty $\rightarrow$ \textbf{Social cognition}
        \item Focus drifting/Wrong topic $\rightarrow$ \textbf{Attention}
        \item Slow/Laggy but retains info $\rightarrow$ \textbf{Processing speed}
        \item Forgetting recent info $\rightarrow$ \textbf{Memory}
        \item \textit{Note:} Stress/Mood do NOT decide the label; base choice on cognitive thinking patterns.
    \end{itemize}
    
    \vspace{4pt}
    \hrule height 0.5pt
    \vspace{4pt}
    
    \textbf{4. Output Format Requirements}\\
    Output a single JSON object. Do NOT include any extra keys or text outside the JSON.
    
    \begin{tcolorbox}[colback=white, colframe=gray!20, boxrule=0.5pt, sharp corners, left=4pt, right=4pt, top=2pt, bottom=2pt]
    \ttfamily\scriptsize
    \{ \\
    \ \ "reflection": "Briefly explain reasoning (2-5 sentences), pointing to key behaviors.", \\
    \ \ "labels": ["Category Name"] \textrm{OR} ["Category A", "Category B"] \\
    \}
    \end{tcolorbox}
    
    \vspace{2pt}
    \textbf{Command:} Now read the conversation and produce your JSON.
    \end{tcolorbox}
    
    \caption{The clinical rater instructions for cognitive domain evaluation.} 
    \label{fig:evaluator_prompt}
\end{figure*}

\begin{figure*}[t!]
    \centering
    \begin{tcolorbox}[
        colback=promptbg, colframe=promptframe, boxrule=0.5pt, arc=2mm,
        title=\textbf{System Prompt for Healthy Control},
        coltitle=black, fonttitle=\small\sffamily, enhanced,
        top=8pt, bottom=8pt, left=10pt, right=10pt
    ]
        \small\sffamily
        \textbf{Role:} You are a cognitively intact, mentally healthy adult client speaking with a counselor. Act strictly in the first person.

        \textbf{Instructions:}
        \begin{itemize}
            \item \textbf{Cognition:} You have NO cognitive deficits. Your memory, attention, processing speed, reasoning, and social understanding are within the normal range.
            \item \textbf{Tone:} Be natural and realistic. Express a normal range of human emotions but remain functional and coherent.
            \item \textbf{Context:} Use the provided background information as your personal history.
        \end{itemize}

        \textbf{Input:} Case Profile: [CASE\_TEXT]
    \end{tcolorbox}
    \caption{System prompt configuration for the Healthy Control baseline.}
    \label{fig:healthy_control_prompt}
\end{figure*}

\begin{figure*}[t!]
    \centering
    \begin{tcolorbox}[
        colback=promptbg, colframe=promptframe, boxrule=0.5pt, arc=2mm,
        title=\textbf{System Prompt for Direct Prompt Baseline},
        coltitle=black, fonttitle=\small\sffamily, enhanced,
        top=8pt, bottom=8pt, left=10pt, right=10pt
    ]
        \small\sffamily
        \textbf{Role:} You are an adult patient talking with a clinical psychologist in an outpatient visit.

        \textbf{Cognitive Condition:}
        You have a mild cognitive difficulty mainly affecting your [COGNITIVE\_DOMAIN].
        \textit{The domain placeholder is dynamically replaced by one of the following natural descriptions:}
        \begin{itemize}
            \item \textbf{Memory}: "memory (you often forget things or lose track of information)"
            \item \textbf{Attention}: "attention (it is hard to stay focused or stay on one thing)"
            \item \textbf{Processing Speed}: "processing speed (you feel slower than before when thinking)"
            \item \textbf{Reasoning \& Problem Solving}: "reasoning and everyday problem solving"
            \item \textbf{Social Cognition}: "social understanding (reading other people and social situations)"
        \end{itemize}

        \textbf{Constraints:}
        \begin{itemize}
            \item Answer strictly as the patient in everyday language (first person).
            \item Keep replies short (1–3 sentences). No bullet points.
            \item Do not act like a doctor or AI.
        \end{itemize}

        \textbf{Input:} Case Profile: [CASE\_TEXT]
    \end{tcolorbox}
    \caption{System prompt configuration for the Direct Prompting baseline, featuring dynamic domain injection.}
    \label{fig:direct_prompt_baseline}
\end{figure*}

\begin{figure*}[t!]
    \centering
    \begin{tcolorbox}[
        colback=promptbg, colframe=promptframe, boxrule=0.5pt, arc=2mm,
        title=\textbf{Prompt for Generating Deficit-Specific CCDs},
        coltitle=black, fonttitle=\small\sffamily, enhanced,
        top=8pt, bottom=8pt, left=10pt, right=10pt
    ]
        \small\sffamily
        \textbf{Role} \\
        You are an expert Clinical Psychologist and Neuropsychologist specializing in Cognitive Behavioral Therapy (CBT) Case Formulation.

        \textbf{Task} \\
        Synthesize a \textbf{Cognitive Conceptualization Diagram (CCD)} based on a provided neutral demographic profile. You must simulate a scenario where this individual is experiencing specific memory deficits that disrupt their daily functioning.

        \textbf{Transformation Logic} \\
        1. Synthesize the input demographics into a clinical summary. \\
        2. Inject a specific, plausible memory deficit that conflicts with the patient's lifestyle or personality. \\
        3. Derive the psychological profile (beliefs, thoughts, behaviors) resulting from this deficit.

        \textbf{Output Specification (JSON Only)} \\
        Return a single JSON object containing the following fields.
        \begin{itemize}
            \item \textbf{name} A short first name.
            \item \textbf{history} A clinical summary including exactly ONE sentence describing a realistic memory deficit.
            \item \textbf{core\_beliefs} Fundamental beliefs activated by the impairment (e.g., 'I am incompetent').
            \item \textbf{intermediate\_beliefs} Conditional rules (e.g., 'If I write everything down, I can hide my failure').
            \item \textbf{coping\_strategies} Strategies used to manage the deficit (e.g., excessive list-making).
            \item \textbf{automatic\_thoughts} Specific thoughts triggered during a moment of failure.
            \item \textbf{emotions} Affective response (e.g., Anxious, Humiliated).
            \item \textbf{behaviors} Observable behavioral reactions.
        \end{itemize}

        \textbf{Input:} Patient Profile: [PATIENT\_INFO]
    \end{tcolorbox}
    \caption{Prompt structure for synthesizing deficit-specific Cognitive Conceptualization Diagrams (CCDs).}
    \label{fig:gen_ccd_prompt}
\end{figure*}

\begin{figure*}[t!]
    \centering
    \begin{tcolorbox}[
        colback=promptbg, colframe=promptframe, boxrule=0.5pt, arc=2mm,
        title=\textbf{System Prompt for User Profile Extraction},
        coltitle=black, fonttitle=\small\sffamily, enhanced,
        top=10pt, bottom=10pt, left=10pt, right=10pt
    ]
        \small\sffamily
        Read the following medical case and write a short English paragraph (50–100 words) summarizing only the patient's basic non-medical information.
        Include a simple, randomly generated English first name that does not appear in the case text. Prefer to use only information explicitly stated in the case for age, gender, family, occupation or studies, living situation, and other non-medical background details.
        If the case contains very little non-medical information, you may add a small number of neutral, non-medical personal details (such as generic occupation, hobbies, or daily life habits) to keep the paragraph between 50 and 100 words. Any invented details must remain clearly non-medical.
        Strictly exclude any mention of physical or mental diseases, symptoms, diagnoses, tests, treatments, surgeries, allergies, medications, family disease history, clinical impressions, or reasons for medical consultation. Do not use bullet points, lists, headings, or meta-commentary. If some types of basic information are missing and you do not wish to invent them, simply omit them.
        
        \textbf{Case:} \{Case information\}
    \end{tcolorbox}
    \caption{System prompt utilized for extracting neutral non-medical user profiles from clinical case reports.}
    \label{fig:profile_extraction_prompt}
\end{figure*}

\begin{figure*}[t!]
    \centering
    \begin{tcolorbox}[
        colback=promptbg, colframe=promptframe, boxrule=0.5pt, arc=2mm,
        title=\textbf{System Prompt for Therapist Agent},
        coltitle=black, fonttitle=\small\sffamily, enhanced,
        top=8pt, bottom=8pt, left=10pt, right=10pt
    ]
        \small\sffamily
        \textbf{Role:} You are a clinical therapist conducting an outpatient consultation.

        \textbf{Task:}
        \begin{itemize}
            \item Understand the patient's emotional and daily life status.
            \item Offer warm, realistic support.
            \item Subtly observe cognitive functions (memory, attention, processing, reasoning \& problem solving, social cognition).
        \end{itemize}

        \textbf{Constraints:}
        \begin{itemize}
            \item NEVER use terms like ``domains'', ``deficits'', or ``test''.
            \item Avoid blunt questions. Use gentle, indirect inquiries.
            \item Keep replies short (1–3 sentences) and natural.
        \end{itemize}

        \textbf{Input:} Patient Profile: [PATIENT\_INFO]
    \end{tcolorbox}
    \caption{System prompt configuration for the conversational Therapist Agent.}
    \label{fig:therapist_agent_prompt}
\end{figure*}

\begin{figure*}[t!]
    \centering
    \begin{tcolorbox}[
        colback=gray!10, colframe=black, boxrule=0.8pt, arc=2mm, 
        title=\textbf{System Prompt for Impairment Severity Consistency (ISC) Evaluation},
        coltitle=white, fonttitle=\small\sffamily\bfseries, enhanced,
        top=8pt, bottom=8pt, left=10pt, right=10pt
    ]
        \small\sffamily
        \textbf{Role:} You are an expert evaluator in computational cognitive modeling and clinical assessment.
        
        \vspace{0.3em} 
        \textbf{Target Domain:} \texttt{[Insert Domain]} 
        
        \vspace{0.3em}
        \textbf{Context:} You are presented with a shuffled triplet of dialogues representing three distinct levels of \textbf{\texttt{[Insert Domain]}} impairment (e.g., \textit{Memory}, \textit{Attention}).
        
        \vspace{0.3em}
        \textbf{Task:} Analyze the symptom intensity in each dialogue based on the clinical criteria for \textbf{\texttt{[Insert Domain]}}. Reconstruct the ground-truth order by ranking them from \textbf{Most Severe} to \textbf{Least Severe}.

        \vspace{0.3em}
        \textbf{Input:}
        \begin{itemize}
            \item \textbf{Dialogue A:} [Insert Dialogue Content]
            \item \textbf{Dialogue B:} [Insert Dialogue Content]
            \item \textbf{Dialogue C:} [Insert Dialogue Content]
        \end{itemize}

        \vspace{0.3em}
        \textbf{Output Format:} Provide your response strictly in valid \textbf{JSON} format without markdown. Use the following structure:
        
        \vspace{0.5em}
        \texttt{\{ \\
        \hspace*{1em} "reasoning": "A brief analysis comparing the symptom intensity...", \\
        \hspace*{1em} "ranking": [ "Dialogue X", "Dialogue Y", "Dialogue Z" ] \\
        \}}
    \end{tcolorbox}
    \caption{The evaluator prompt template used to calculate Impairment Severity Consistency (ISC). The placeholder \texttt{[Insert Domain]} is dynamically replaced with the specific cognitive domain being evaluated (e.g., Memory, Social Cognition). The model is tasked with recovering the correct severity order (\textit{Severe} $>$ \textit{Moderate} $>$ \textit{Mild}).}
    \label{fig:isc_eval_prompt}
\end{figure*}

\end{document}